\documentclass[journal]{IEEEtran}
\usepackage{float}

\usepackage{booktabs}
\usepackage{multirow}
\usepackage{amsmath}

\usepackage{makecell}

\usepackage{graphicx}
\usepackage{enumitem}

\usepackage{graphicx}
\usepackage{subfig}  

\usepackage{xcolor}
\usepackage{hyperref}

\hypersetup{
    colorlinks=true,
    linkcolor=blue,
    citecolor=blue,
}

\usepackage{caption}
\usepackage{array} 
\newcolumntype{P}[1]{>{\raggedright\arraybackslash}p{#1}}

\ifCLASSINFOpdf
\else
\fi
\begin{document}
%
\title{DI3CL: Contrastive Learning With Dynamic Instances and Contour Consistency for SAR Land-Cover Classification Foundation Model }
%
%
%

\author{ Zhongle~Ren,~\IEEEmembership{Member,~IEEE},
        Hui~Ding,
        Kai~Wang,
        Biao~Hou,~\IEEEmembership{Senior Member,~IEEE},
        Xingyu~Luo,
        Weibin~Li,~\IEEEmembership{Member,~IEEE},
and~Licheng~Jiao,~\IEEEmembership{Fellow,~IEEE}

\thanks{This work was supported in part by the National Natural Science Foundation of China under Grant,62171347,62401418,62371373,62271377. The Key Research and Development Program in Shaanxi Province of China under Grant 2024CY2-GJHX-17.
The Key Industry Chain Technology Research and Development General Project of Xi'an under Grant 2024JH-CLYB-0047. \textit{(Co-corresponding author: Kai Wang and Biao Hou.)}}
\thanks{The authors are with the Key Laboratory of Intelligent Perception and Image Understanding of Ministry of Education, International Research Center for Intelligent Perception and Computation, Joint International Research Laboratory of Intelligent Perception and Computation, 
School of Artificial Intelligence, Xidian University, Xi’an 710071
(email:zlren@xidian.edu.cn;18250107162@163.com;kiwi@stu.xidian.edu.cn;
avcodec@hotmail.com;luoxingyustu@163.com;weibinli@xidian.edu.cn; lchjiao@mail.xidian.edu.cn). Weibin Li is also with Hangzhou Institute of Technology.
This paper has supplementary downloadable material available at http://ieeexplore.ieee.org., provided by the author. The material includes supplemental material. Contact kiwi@stu.xidian.edu.cn for further questions about this work.
}}

%
%

\markboth{Journal of \LaTeX\ Class Files,~Vol.~3, No.~6, June~2025}%
{Shell \MakeLowercase{\textit{et al.}}: Bare Demo of IEEEtran.cls for Journals}
%



\maketitle

\begin{abstract}
Although significant advances have been achieved in SAR land-cover classification, recent methods remain predominantly focused on supervised learning, which relies heavily on extensive labeled datasets. This dependency not only limits scalability and generalization but also restricts adaptability to diverse application scenarios. In this paper, a general-purpose foundation model for SAR land-cover classification is developed, serving as a robust cornerstone to accelerate the development and deployment of various downstream models. Specifically, a Dynamic Instance and Contour Consistency Contrastive Learning (DI3CL) pre-training framework is presented, which incorporates a Dynamic Instance (DI) module and a Contour Consistency (CC) module. DI module enhances global contextual awareness by enforcing local consistency across different views of the same region. CC module leverages shallow feature maps to guide the model to focus on the geometric contours of SAR land-cover objects, thereby improving structural discrimination. Additionally, to enhance robustness and generalization during pre-training, a large-scale and diverse dataset named SARSense, comprising 460,532 SAR images, is constructed to enable the model to capture comprehensive and representative features. To evaluate the generalization capability of our foundation model, we conducted extensive experiments across a variety of SAR land-cover classification tasks, including SAR land-cover mapping, water detection, and road extraction. The results consistently demonstrate that the proposed DI3CL outperforms existing methods. Our code and pre-trained weights are publicly available at: https://github.com/SARpre-train/DI3CL.
\end{abstract}


%
\IEEEpeerreviewmaketitle

\section{Introduction}
\IEEEPARstart{S}{ynthetic} Aperture Radar (SAR) is a cutting-edge active microwave sensing technology capable of capturing data under all-day and all-weather conditions \textcolor{blue}{\cite{all_day1}}, \textcolor{blue}{\cite{all_day2}}. This ability to acquire imagery regardless of lighting or atmospheric factors, such as fog, clouds, and rain, makes SAR a powerful tool in both civil and military fields \textcolor{blue}{\cite{civil}}, \textcolor{blue}{\cite{military}}. With the growing number of satellite missions, the volume of SAR images has expanded significantly, leading to an increasing need for automated methods to interpret these images. Land-cover classification, which assigns semantic labels to each pixel, is crucial for understanding SAR data and has wide applications in areas such as environmental monitoring \textcolor{blue}{\cite{monitoring,monitoring2,monitoring3}}, agriculture \textcolor{blue}{\cite{agriculture,agriculture2}}, and urban development \textcolor{blue}{\cite{development,development2}}.

Researchers have proposed various approaches for SAR land-cover classification, which can be broadly classified into two categories: conventional approaches and data-driven Deep Learning (DL) approaches. Conventional approaches to SAR interpretation involve extracting handcrafted features based on intrinsic characteristics of the data \textcolor{blue}{\cite{categories}}. These methods can be categorized into physical model-based methods \textcolor{blue}{\cite{HA}}, \textcolor{blue}{\cite{freeman}}, \textcolor{blue}{\cite{pauli}} and statistical model-based methods \textcolor{blue}{\cite{statistical method1}}, \textcolor{blue}{\cite{statistical method2}}, \textcolor{blue}{\cite{statistical method3}}. Although model-based approaches offer better interpretability, they often require time-consuming, scenario-specific feature selection and classifier design. Additionally, issues such as speckle noise and the unique geometric characteristics of land-cover objects in SAR imagery can compromise robustness, limiting their effectiveness in practical scenarios \textcolor{blue}{\cite{drawbacks}}. In contrast, data-driven DL approaches can process diverse data types, hierarchically extract features, and automatically predict land-cover categories, offering significant advantages over model-based SAR classification approaches. Therefore, there is growing interest in developing DL methods for SAR land-cover classification. Among them, fully convolutional network-based approaches, such as FCN \textcolor{blue}{\cite{FCN}}, U-Net \textcolor{blue}{\cite{UNet}},\textcolor{blue}{\cite{DeepAqua}},\textcolor{blue}{\cite{WVResU-Net}}, DeepLab \textcolor{blue}{\cite{Deeplabv1}}, \textcolor{blue}{\cite{Deeplabv2}}, \textcolor{blue}{\cite{Deeplabv3}}, have become dominant in this field. For instance, existing methods have proposed specialized DL architectures for SAR classification, such as noise-mitigating convolution-wavelet networks and FCNs tailored for polarimetric SAR wetland classification \textcolor{blue}{\cite{CWNN}}, \textcolor{blue}{\cite{SAR2}}. While these methods have demonstrated promising results, they depend significantly on extensive manually labeled training datasets for supervised learning. On the one hand, since SAR land-cover classification operates at the pixel level and is further challenged by speckle noise, the annotation process often requires expert knowledge, resulting in high costs and significant time investment. On the other hand, the datasets obtained are often limited in applicability, necessitating repeated iterations in data collection and model development. This iterative process not only significantly increases cumulative time and financial costs but also hinders the applicability and scalability of DL frameworks for SAR land-cover classification \textcolor{blue}{\cite{SMLFR}}.

    

\begin{figure}[t]
    \centering
    \includegraphics[width=0.48\textwidth]{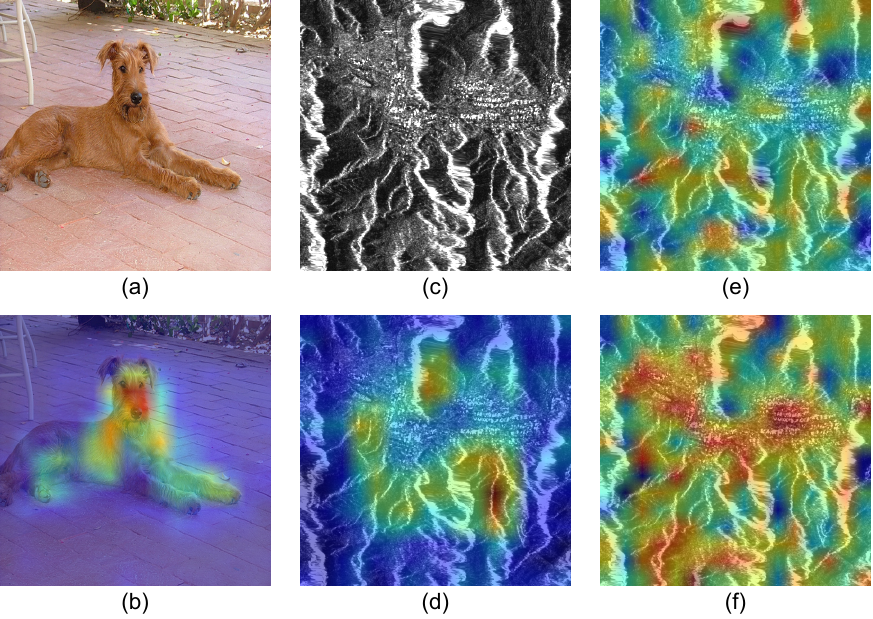} 
    \caption{Illustration of issues with existing self-supervised CL for SAR land-cover classification and our improvements. Visualizations use Attention-Gradient Class Activation Mapping (A-GCAM)~\cite{AGCAM}. (a)-(b): For natural images, the model focuses on foregrounds. (c)-(d): For SAR images, it only attends to central regions (e.g., village) and ignores others (e.g., forest). (e): Adding Dynamic Instances expands focus to the whole image. (f): Including Contour Consistency ensures the model attends to the full image and preserves land-cover contours.}
    \label{fig_motivation}
\end{figure}

Recently, foundation models \textcolor{blue}{\cite{foundation-model}} using self-supervised learning \textcolor{blue}{\cite{self-supervised}} have advanced computer vision by enabling cross-task knowledge transfer with minimal labeled data, inspiring remote sensing models \textcolor{blue}{\cite{RSP}}, \textcolor{blue}{\cite{ringmo}}, \textcolor{blue}{\cite{SatMAE}}, \textcolor{blue}{\cite{Scale-MAE}}, \textcolor{blue}{\cite{RingMoE}}. However, most are designed for optical images and underperform on SAR data due to significant domain gaps, such as speckle noise and unique scattering characteristics. There are a few SAR-specific models \textcolor{blue}{\cite{SAR-JEPA}}, \textcolor{blue}{\cite{MSFA}}, \textcolor{blue}{\cite{SARATR-X}}, \textcolor{blue}{\cite{of_add}}, but they only address object-level tasks and lack pixel-level details. Multi-source models \textcolor{blue}{\cite{SkySense}},  \textcolor{blue}{\cite{OFA-Net}}, \textcolor{blue}{\cite{RingMoE}} operate cross-modally but tend to be inefficient and overlook specific features of land-cover objects in SAR images. Consequently, the field still lacks a SAR foundation model tailored for pixel-level segmentation that enables rapid, label-efficient development and deployment. This gap motivates us to make the first attempt to construct a SAR land-cover classification foundation model, which serves as a robust cornerstone for efficient downstream model implementation. Currently, there are two mainstream approaches: Vision Transformers (ViTs) based on Masked Image Modeling (MIM) \textcolor{blue}{\cite{RVSA}}, and Convolutional Neural Networks (CNNs) based on Contrastive Learning (CL) \textcolor{blue}{\cite{pis}}. MIM-based methods are designed to reconstruct occluded pixels and learn general feature representations from data distributions. However, they are not well-suited for pixel-level SAR land-cover classification, as the inherent speckle noise in SAR images poses significant challenges for accurate image reconstruction \textcolor{blue}{\cite{fgmae}}. Furthermore, the key information in SAR images lies in shapes and geometric structures rather than pixel-level textures, limiting the effectiveness of MIM in capturing these critical aspects. In contrast, CL-based methods utilize an instance discrimination pretext task, training the model to maximize similarity between different augmented views of the same instance while minimizing similarity to other instances, thereby learning more discriminative and robust representations \textcolor{blue}{\cite{instance1}}, \textcolor{blue}{\cite{instance2}}. However, most existing CL methods are specifically designed for natural images, which typically contain clear foreground instances and background. In these scenarios, the instance discrimination task leads the model to focus on the individual instances, as illustrated in Fig.\ref{fig_motivation}(b). However, the SAR images captured from high altitudes do not exhibit well-defined instances. As a result, directly applying CL to SAR images tends to guide the model to focus only on the central regions of the image. This not only leads to the loss of critical information in non-central areas of SAR data with heterogeneous backscattering but also causes the model to overlook the crucial shapes and geometric structures inherent to SAR images, severely impairing the performance of land-cover classification. Fig.~\ref{fig_motivation}(d) illustrates that the model attends exclusively to the central village while ignoring the surrounding forested areas. Directly applying instance discrimination tasks-based CL to SAR land-cover classification is therefore ineffective, as such tasks require global context perception and adequate extraction of land-cover features in SAR images.

In summary, while DL-based approaches have been the leading approach for SAR land-cover classification, they remain constrained by their dependence on extensive labeled datasets, which are hard to obtain due to both cost and complications from speckle noise and the absence of clear color or texture cues that assist feature learning.  Although CL-based foundation models help reduce reliance on labels and speed up model development, they still struggle to extract robust features from SAR images, as they tend to focus only on the central regions, which is especially problematic given the indistinct foreground-background boundaries inherent in SAR imagery. To overcome these limitations, the Dynamic Instance and Contour Consistency Contrastive Learning (DI3CL) pre-training framework is proposed, a dedicated architecture for SAR land-cover classification that integrates two custom modules to address the aforementioned SAR-specific challenges. This framework ensures the pre-trained model captures the global context of SAR images while aligning focus across land-cover objects, as depicted in Fig.\ref{fig_motivation}(f). Specifically, to resolve the center bias issue exacerbated by indistinct foreground-background boundaries, scattered land-cover distributions, and homogeneous textures in SAR imagery, the Dynamic Instance (DI) module is introduced to improve the perception of local details and enhance global contextual understanding. As shown in Fig.\ref{fig_motivation}(e), DI enables the model to attend to the entire image. Additionally, to address the lack of color or texture cues in SAR imagery, where shapes and geometric structures serve as key distinguishing features, the Contour Consistency (CC) module is introduced. This module leverages shallow features abundant in contour information to enable instance discrimination, thereby directing the model’s attention to the shapes and geometric structures of land cover in SAR imagery, enabling superior feature extraction even in the presence of speckle noise. To further advance the generalization capabilities of the base model, alleviate the labeled-data bottleneck in DL for SAR, and foster progress in SAR feature representation learning, a large-scale dataset comprising over 460,532 SAR images, named SARSense, is curated. SARSense offers diverse and representative samples spanning multi-source, multi-resolution, and multi-polarization SAR data, providing a robust foundation for feature learning and propelling the advancement of research for the SAR foundation model.

The main contributions of this article are summarized as follows.
\begin{enumerate}
    \item The DI3CL pre-training framework leverages the synergistic integration of DI and CC modules to address the challenges of SAR image land-cover classification. DI reduces center bias, while CC enhances sensitivity to land-cover shapes and structures, even in SAR images with speckle noise. This collaboration optimizes feature extraction and boosts classification performance.
    \item The SARSense dataset is constructed, comprising over 460,532 multi-source, multi-resolution, and multi-polarization SAR images. It provides abundant, representative training samples to alleviate the labeled-data bottleneck, enhance the model’s SAR-specific visual representation capabilities, and further propel the development of SAR feature representation learning.
    \item The first publicly available foundation model for SAR land-cover classification is proposed, achieving state-of-the-art performance across various downstream tasks, including SAR land-cover mapping, water detection, and road extraction. This demonstrates the model’s robustness and generalization in practical scenes, while lowering the threshold for downstream pixel-level SAR tasks.
\end{enumerate}

\begin{table*}[t]
\setcounter{table}{0}
\centering
\caption{Detailed statistics of the SARSence dataset}
\begin{tabular}{c c c c c c c }
\hline
\textbf{Dataset} & \textbf{Source} & \textbf{Land-Cover Types} & \textbf{Spatial Resolution} & \textbf{Polarization} & \textbf{Size} & \textbf{Scale} \\
\hline
SARSence & Gaofen-3, Sentinel-1 & \parbox{3cm}{\vspace{1mm} water, farmland, meadow, \\ village, forest, road, etc \vspace{1mm}} & 1,3,5,8,10,25m & HH,HV,VH,VV & 512 × 512 & 460,532 \\
\hline
\end{tabular}
\label{tab:sar_datasets}
\end{table*}

\begin{figure*}[htbp]
    \centering
    \includegraphics[width=1.00\textwidth]{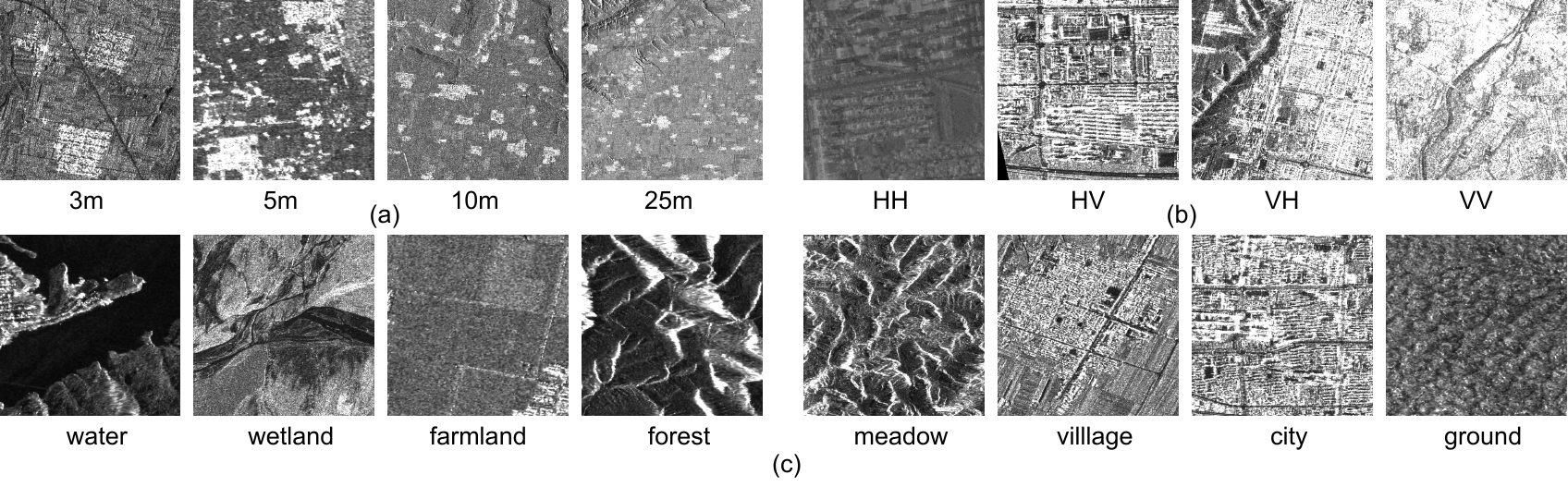} 
    \caption{Illustrative samples from the SARSense dataset: (a) samples of village and farmland under different spatial resolutions, demonstrating the diversity of resolution; (b) samples of city areas acquired using different polarization modes, demonstrating the diversity of polarization; (c) samples of various land-cover categories, demonstrating the diversity of land-cover types.}
    \label{fig:figure_dataset}
\end{figure*}

\section{RELATED WORKS}
\subsection{Deep Learning in SAR Land-Cover Classification}

With the advancement of DL techniques, SAR land-cover classification has steadily shifted from traditional methods to DL approaches. CWNN  \textcolor{blue}{\cite{CWNN}} introduces a convolution-wavelet architecture that effectively reduces speckle noise while preserving structural features in SAR images. Mohammadimanesh et al.  \textcolor{blue}{\cite{SAR2}} design a novel FCN architecture specifically tailored for classifying wetland complexes using Polarimetric SAR (PolSAR).
RCC-MRF \textcolor{blue}{\cite{RCC-MRF}} integrates CNN-extracted deep features with region-level spatial constraints, using a region category confidence mechanism and Markov random field to suppress both pixel- and region-level misclassifications. Liang et al.  \textcolor{blue}{\cite{HCRF}} advance pixel-wise classification in high-resolution SAR imagery by integrating a context-aware encoder network with a hybrid conditional random field (HCRF), effectively capturing both local and global context to enhance label consistency. CCNR  \textcolor{blue}{\cite{CCNR}} introduces a cross-regional context and noise regularization strategy, leveraging CL and self-attention to enhance robustness against terrain variations and image noise. DeepAqua \cite{DeepAqua} proposes an unsupervised knowledge distillation framework, using NDWI-derived open-water masks from optical imagery as a teacher to train a U-Net student model on SAR data, enabling accurate segmentation of both open and vegetated wetland water without any manual annotations.
WVResU-Net \cite{WVResU-Net} proposes a vision MLP-enhanced ResU-Net with residual connections and wave-based token processing, achieving superior flood mapping performance on dual-polarization Sentinel-1 SAR imagery compared to existing CNN and transformer baselines.

Meanwhile, the fusion of SAR and optical data for land-cover classification has become a growing trend, owing to their complementary characteristics. SAR images provide structure-preserving, weather-insensitive information, and optical images offer fine spectral details. For instance, MCANet  \textcolor{blue}{\cite{MCANet}} adopts a pseudo-siamese structure with cross-attention and multi-scale feature fusion to jointly process optical and SAR images, leading to increased classification accuracy across various land-cover types. Li et al.  \textcolor{blue}{\cite{circular module}} further address semantic misalignment through a spatial-aware circular module and latent feature space alignment, achieving better fusion and classification outcomes. SoftFormer  \textcolor{blue}{\cite{SoftFormer}} proposes a multi-level fusion transformer that integrates CNNs and transformers using a joint key–value learning mechanism, outperforming state-of-the-art models on multiple urban land-use classification tasks, even under cloud-covered conditions. These fusion methods highlight the importance of spatial context and semantic alignment, which also guide our design of SAR-only modules to achieve similar goals without relying on optical data.

Despite notable advancements, DL models for SAR land-cover classification predominantly rely on extensive, manually annotated datasets. Given the pixel-level granularity and the necessity for expert knowledge in labeling, this process is both costly and time-consuming. Furthermore, the limited generalizability of existing datasets often necessitates repeated data collection and model retraining across different regions, thereby escalating both time and financial expenditures. These challenges significantly impede the scalability and practical deployment of DL approaches in SAR land-cover classification.

\subsection{Self-Supervised Learning}
To alleviate the challenge of annotation scarcity, SSL has been introduced to extract meaningful feature representations from unlabeled data and enhance performance in downstream tasks. SSL methods can be categorized into three main types: predictive, generative, and contrastive approaches.

Predictive methods aim to learn visual representations by solving pretext tasks that involve predicting inherent structures or transformations, including, but not limited to, position prediction \textcolor{blue}{\cite{position}}, jigsaw puzzles \textcolor{blue}{\cite{jigsaw}}, and inpainting \textcolor{blue}{\cite{inpainting}}. In the context of SAR imagery, JPSSL  \textcolor{blue}{\cite{JPSSL}} employs a jigsaw puzzle task to learn semantic features from SAR images, achieving high land-cover classification accuracy with minimal labeled data. RotANet  \textcolor{blue}{\cite{RotANet}} focuses on rotation prediction among intra-class SAR target images to enhance generalization performance under limited data conditions.


Generative methods are designed to reconstruct inputs or to predict latent structures to learn robust representations. Representative frameworks include MAE \textcolor{blue}{\cite{MAE}} and SimMIM \textcolor{blue}{\cite{SimMIM}}, which are typical of MIM-based approaches extended to the SAR image domain. Such MIM methods primarily focus on SAR target recognition tasks. S2FL \textcolor{blue}{\cite{S2FL}} proposes a multimodal generative framework leveraging complementary views for land-cover classification. 

Contrastive methods learn representations by distinguishing between similar and dissimilar samples. In computer vision, this paradigm began with instance discrimination frameworks such as MoCo \textcolor{blue}{\cite{moco}} and SimCLR \textcolor{blue}{\cite{simclr}}, which learn by contrasting positive and negative sample pairs. MoCo introduced a momentum encoder to maintain consistent representations, while SimCLR highlighted the importance of strong data augmentations and larger batch sizes. Later, BYOL \textcolor{blue}{\cite{byol}} and SimSiam \textcolor{blue}{\cite{simsiam}} removed the need for negative samples by leveraging asymmetric network structures and stop-gradient mechanisms. Barlow Twins \textcolor{blue}{\cite{barlowtwins}} further advanced this trend by decorrelating feature dimensions to reduce redundancy without requiring negative pairs. With the rise of Vision Transformers (ViTs), methods like MoCov3 \textcolor{blue}{\cite{mocov3}}, DINO \textcolor{blue}{\cite{dino}}, and DINOv2 \textcolor{blue}{\cite{DINOv2}}extended CL to token-based representations, enabling better global semantic understanding. Since CL typically focuses on instance-level features, many CL for Remote Sensing (RS) approaches emphasize the extraction of local features. GLCNet  \textcolor{blue}{\cite{GLCNet}} combines global style and local patch contrast for semantic segmentation. IndexNet  \textcolor{blue}{\cite{IndexNet}} tracks spatial positions for pixel-level representation learning. SSCCL  \textcolor{blue}{\cite{ssccl}} integrates instance-level and region-level consistency branches to enhance both spatial and semantic representation quality. Furthermore, in the SAR image domain, CSSL  \textcolor{blue}{\cite{CSSL}} adopts a coarse-to-fine strategy to progressively learn global and local features.

These methods constitute a critical basis for building foundation models in remote sensing and serve as early yet significant efforts in this emerging research direction.

\subsection{Foundation Models in Remote Sensing}
RS-oriented Self-Supervised Learning (SSL), combined with large-scale datasets \textcolor{blue}{\cite{RSP}} and massive model parameters \textcolor{blue}{\cite{a billion scale}}, has facilitated the development of foundation models in RS. The construction of these models primarily relies on two approaches: Masked Image Modeling (MIM) and Contrastive Learning (CL). However, existing RS foundation models are predominantly designed for optical imagery or multimodal data fusion, with a notable lack of dedicated efforts on single-source SAR foundation models tailored for the core land-cover classification task of SAR images.

MIM has emerged as a mainstream method for building RS  foundation models. SatMAE \textcolor{blue}{\cite{SatMAE}} leverages temporal and multi-spectral satellite imagery through a masked autoencoder framework, achieving significant improvements in supervised and transfer learning tasks. RVSA \textcolor{blue}{\cite{RVSA}} advances plain Vision Transformers with rotated varied-size window attention to handle large-scale RS images, demonstrating superior performance in detection tasks. RingMo \textcolor{blue}{\cite{ringmo}} introduces a generative SSL framework tailored for dense and small objects in complex scenes, achieving state-of-the-art results across multiple datasets. Scale-MAE \textcolor{blue}{\cite{Scale-MAE}} explicitly learns multiscale relationships by reconstructing low- and high-frequency images at different scales, enhancing geospatial representation. SMLFR \textcolor{blue}{\cite{SMLFR}} designs a generative ConvNet foundation model with sparse modeling and low-frequency reconstruction, optimizing performance for dense prediction tasks. SARATR-X \textcolor{blue}{\cite{SARATR-X}}, SAR-JEPA \textcolor{blue}{\cite{SAR-JEPA}}, MSFA \textcolor{blue}{\cite{MSFA}}, and SPT \textcolor{blue}{\cite{of_add}} specialize in SAR target recognition, employing self-supervised learning to effectively generalize across various tasks with minimal labeled data, though they do not tackle pixel-level land-cover classification. Multi-source models such as SkySense \textcolor{blue}{\cite{SkySense}}, OFA-Net \textcolor{blue}{\cite{OFA-Net}}, SkySenseV2 \textcolor{blue}{\cite{SkySenseV2}}, and RingMoE \textcolor{blue}{\cite{RingMoE}} integrate multimodal RS data and excel in optical-SAR scenarios, but remain limited for SAR-only land-cover classification, where dedicated SAR foundation models offer clearer advantages.

CL has also been widely adopted for constructing remote sensing foundation models. Ayush et al. \textcolor{blue}{\cite{Geography-Aware SSL}} exploit spatio-temporal structures in geo-located datasets to close the gap between contrastive and supervised learning. Seasonal Contrast (SeCo) \textcolor{blue}{\cite{SeCo}} leverages uncurated remote sensing data with time and position invariance to learn transferable representations. PIS \textcolor{blue}{\cite{PIS}} promotes intra-instance similarity by leveraging temporal information, enhancing the model's ability to extract general and invariant features for downstream tasks.

Combining MIM with CL presents another viable approach. CMID \textcolor{blue}{\cite{CMID}} unifies CL and MIM in a self-distillation framework, enabling the learning of both global semantic and local spatial representations for diverse remote sensing tasks.

These advancements highlight the transformative potential of foundation models in RS, driven by innovative SSL techniques and scalable architectures. However, there is a critical gap: dedicated single-source SAR foundation models for key pixel-level tasks, like land-cover classification, are lacking. To address this, we propose a self-supervised CL framework tailored for SAR imagery and develop a SAR land-cover foundation model by pre-training on our large-scale SAR dataset. This approach reduces reliance on labeled data and boosts downstream task performance.

\section{PROPOSED METHOD}
This section introduces the proposed foundation model development framework for SAR land-cover classification, referred to as DI3CL. The framework is designed to train a foundation model from collected SAR data using a contrastive SSL approach. In this section, the SAR dataset we constructed is first presented and is named SARSense. Then, a detailed description of the DI3CL pre-training framework is provided, which incorporates several enhancements over conventional CL methods.

\subsection{SARSense Dataset for Pre-training}
A large-scale and diverse unlabeled SAR pre-training dataset, named SARSense, has been developed. The detailed description of it is presented in Table \ref{tab:sar_datasets}. This dataset comprises SAR imagery collected from sensors with varying imaging parameters, such as Gaofen-3 and Sentinel-1, and includes a variety of polarization modes, including HH, HV, VH, and VV. The inclusion of multi-source and multi-polarization data significantly enhances the robustness and generalization capabilities of models under diverse radar imaging conditions.

SARSense offers extensive geographic and scene diversity, spanning both northern and southern regions of China, including provinces such as Shaanxi and Hubei, as well as international areas such as Berlin and Munich. These areas differ markedly in geography, climate, land-cover classes, and socioeconomic activities, providing a valuable range of scenes for model training and evaluation. This diversity is essential for improving cross-regional generalization in SAR land-cover classification and strengthening SAR feature representation learning. The dataset includes a wide array of land-cover classes, such as urban, farmland, village, water, forest, road, and wetland, which closely match the core categories of SAR land-cover classification. This breadth ensures that pre-trained models can learn distinctive features of a wide range of land covers, promoting adaptability and robust performance across different land-cover classes and establishing a strong basis for generalizable SAR feature learning. To ensure consistency in the pre-training process, all images are uniformly cropped to $512\times512$ pixels, resulting in 460,532 training samples. SARSense spans a wide range of spatial resolutions, with the finest resolution of 1 meter and the coarsest of 25 meters. This multi-resolution characteristic adapts to the diverse demands of SAR land-cover classification, enabling models to acquire scale-invariant features. To our knowledge, SARSense is the largest publicly available SAR pre-training dataset, playing a pivotal role in the development of advanced foundation models for SAR land-cover classification. It addresses the lack of large-scale, diverse SAR pre-training resources, thereby driving progress in SAR feature representation learning and enhancing performance in a variety of downstream SAR tasks. Fig. \ref{fig:figure_dataset} showcases sample selections, underlining the dataset's richness in scene and land-cover diversity.

\begin{figure*}[htbp]
    \centering
    \includegraphics[width=1.0\textwidth]{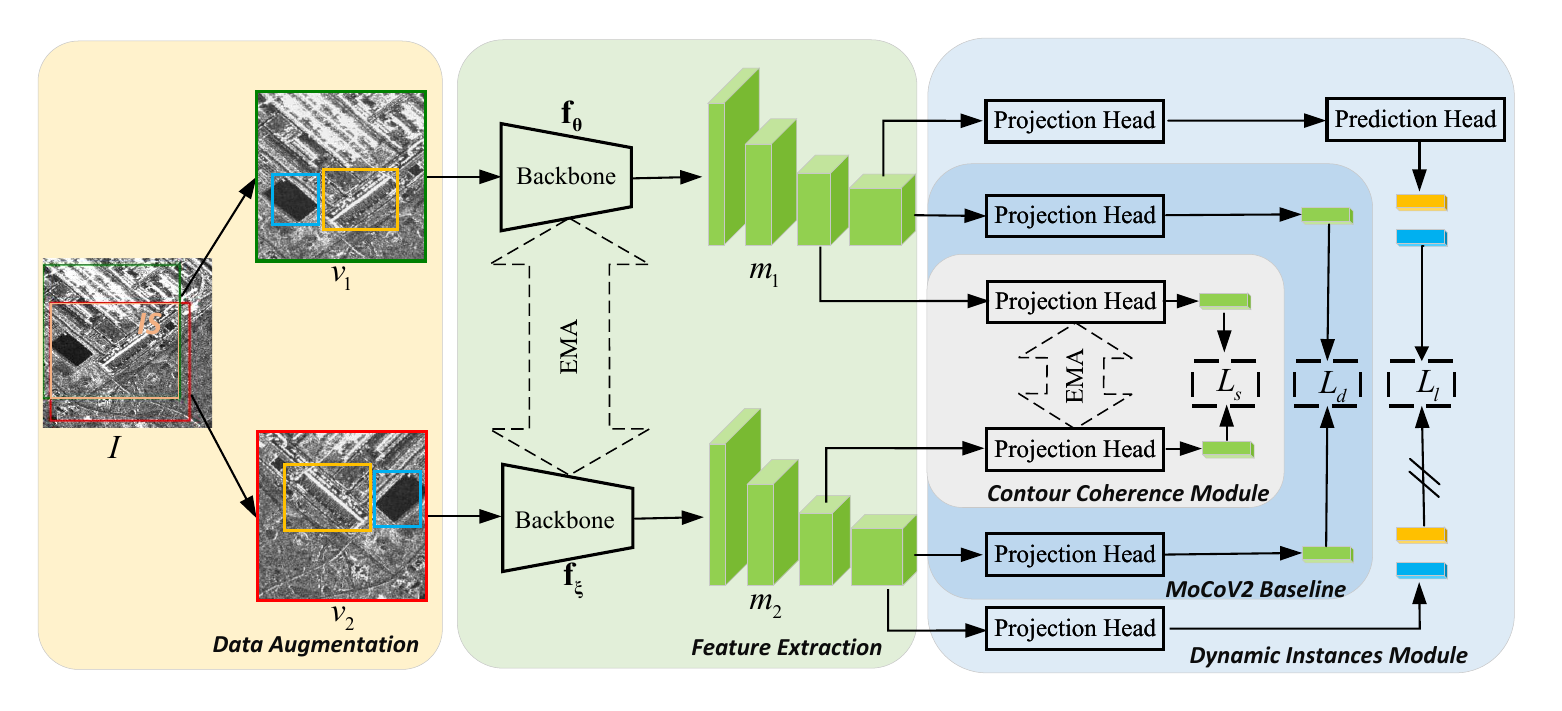} 
    \caption{The overall workflow of the proposed DI3CL framework, where IS denotes the intersection region shared by the two augmented views.}
    \label{fig:figure 0}
\end{figure*}

\subsection{DI3CL Framework}

As illustrated in Fig. \ref{fig:figure 0}, DI3CL is proposed to address the aforementioned challenges specific to SAR land-cover classification and to capture robust feature representations from the SARSense dataset. MoCoV2 \textcolor{blue}{\cite{mocov2}} serves as the baseline due to its established effectiveness in self-supervised learning for high-quality feature extraction. DI3CL is specifically designed to accommodate the unique characteristics of SAR imagery.

DI3CL incorporates two neural networks: an online network, parameterized by $\theta$, and a target network, parameterized by $\xi$. The target network acts as the regression reference for training the online network, with its parameters $\xi$ updated through an exponential moving average of the online network's parameters $\theta$, controlled by a decay factor $m$. This update rule is expressed as:
\begin{equation}
    \xi \leftarrow m\xi + (1 - m)\theta
\end{equation}

Given an unlabeled image $I$ from the pre-training dataset, two separate data augmentation transformations $\mathbf{t_1}$ and $\mathbf{t_2}$ are randomly sampled from the same set of augmentations $\mathbf{T}$. The two sampled transformations are then applied to the image $I$, resulting in two augmented views: $v_1$ and $v_2$. This process can be expressed mathematically as:
\begin{equation}
    v_1 = \mathbf{t_1}(I), v_2 = \mathbf{t_2}(I), \mathbf{t_1},\mathbf{t_2} \in \mathbf{T}
\end{equation}

After obtaining two augmented views, they are passed through the online network and the target network to extract features for subsequent representation learning. Formally, let $(\mathbf{f_{\theta},f_{\xi}})$ represent corresponding network mapping functions. Given the two augmented views $v_1$ and $v_2$, the feature maps extracted by $(\mathbf{f_{\theta},f_{\xi}})$ can be expressed as follows:
\begin{equation}
    m_1^{d} = \mathbf{f_\theta}(v_1), m_2^{d} = \mathbf{f_\xi}(v_2)
\end{equation}
where  $m_1^{d}$ and $m_2^{d}$ denote the deep level feature maps extracted from $v_1$ and $v_2$, respectively. For the sake of notational clarity, the superscript $d$ is employed to distinguish this expression from a similar one to be introduced later in the contour consistency module.

In the MoCoV2 baseline, after obtaining the feature maps, Global Average Pooling (GAP) is applied to obtain the instance-level features from the images. Formally, we express them as $g_1^{d}$ and $g_2^{d}$. After which, a projection head is used to map the extracted instance-level features into a high-dimensional feature representation space, noted as $p_1^{d}$ and $p_2^{d}$. 

Then, MoCoV2 performs an instance discrimination task on the representation vectors. Specifically, the augmented versions of the same sample form a positive pair, while other samples serve as negative samples. The network is tasked with pulling the positive pair closer in the embedding space while pushing the negative samples further apart. MoCoV2 sets up a queue that stores $M$ negative samples, denoted as $\{k_0, k_1, \dots, k_M\}$. In each iteration, $N$ negative samples are updated in the queue based on a ``first-in, first-out" principle.

Finally, the InfoNCE loss is used to optimize this task, which encourages the network to assign higher similarity scores to positive samples while minimizing the similarity scores for negative samples. InfoNCE loss is defined as follows:
\begin{equation}
L_d=-\log \frac{\exp \left(p_1^d \cdot p_2^d / \tau\right)}{\exp \left(p_1^d \cdot p_2^d / \tau\right)+\sum_{i=0}^M \exp \left(p_1^d \cdot k_i^d / \tau\right)}
\end{equation}
where $\tau > 0$ is a temperature coefficient.

It has been found that directly applying MoCoV2 to SAR images leads the model to overly focus on the image center, thereby limiting its ability to perceive global context. To address this, a Dynamic Instance module is proposed. Additionally, to enhance the attention of the model to shape and geometric structures in SAR images, a Contour Coherence module is introduced.

\begin{table*}[!htbp]
    \centering
    \setcounter{table}{1} 
    \footnotesize 
    \setlength{\tabcolsep}{3.5pt} 
    \renewcommand{\arraystretch}{1.3} 
    
    \caption{Detailed Specifications of SARSense Datasets and Downstream Tasks}
    \label{tab:sarsense_fixed}
    
    \begin{tabular}{
        P{2.2cm} 
        P{2.8cm} 
        P{1.2cm} 
        P{1.2cm} 
        P{3.0cm} 
        P{4.8cm} 
    }
    \toprule
    \textbf{Dataset or Task} & \textbf{Region} & \textbf{Res.} & \textbf{Pol.} & \textbf{Scale} & \textbf{Description} \\ \midrule

    SARSense & 
    Extensive geographic diversity, spanning northern and southern China (Shaanxi, Hubei) and international cities (Berlin, Munich). & 
    1, 3, 5, 8, 10, 25m & 
    HH, HV, VH, VV & 
    460,532 images & 
    Massive diverse data collected to provide a rich foundation for large-scale pre-training. \\ \addlinespace

    SAR land-cover pre-train & 
    Randomly selected from the SARSense dataset. & 
    1, 3, 5, 8, 10, 25m & 
    HH, HV, VH, VV & 
    60,000 images & 
    Small-scale set for rapid method validation and hyperparameter tuning during development. \\ \addlinespace

    SAR land-cover downstream & 
    Randomly selected from the SARSense dataset. & 
    3m & 
    HH & 
    10,000 images (Split: Train 30\%, Val 10\%, Test 60\%) & 
    Performance evaluation for models pre-trained on the 60,000-image dataset. \\ \addlinespace

    SAR land-cover mapping & 
    Randomly selected from the SARSense dataset. & 
    3m & 
    HH & 
    33,108 images (Split: Train 80\%, Val 10\%, Test 10\%) & 
    To compare foundation models' performance in all-element land-cover scenarios. \\ \addlinespace

    SAR water detection & 
    Train/Val: Shaanxi, Jiangxi, Anhui. Test: Hunan, Hubei, Henan. & 
    5m & 
    HH & 
    Train/Val: 9,819/1,091 images from 11 scenes. Test: 6,593 images from 4 scenes. & 
    Evaluates water detection. The cross-regional sets fully demonstrate generalization performance. \\ \addlinespace

    SAR road extraction & 
    Guanzhong region, Shaanxi Province (dense road network). & 
    3m & 
    HH & 
    3 scenes, 1,107 tiles (8:1:1 split); Training set augmented to 5,316 images. & 
    Used to compare road extraction capabilities of foundation models. \\
    \bottomrule
    \end{tabular}
\end{table*}
\subsubsection{Dynamic Instances Module}
Inspired by SCRL \textcolor{blue}{\cite{scrl}}, the DI module is designed to enhance local perception throughout the entire SAR image. It addresses center bias, enabling the network to accommodate speckle noise and irregular land-cover patterns in diverse scenarios while also enhancing global contextual awareness. As shown in Fig.\ref{fig:figure 0}, the intersection region of $v_1$ and $v_2$, denoted as $IS(v_1,v_2)$, is firstly identified. Within this intersection, a box $B = (x, y, w, h)$ is randomly sampled. The coordinates of $B$ are then converted to the corresponding positions in each view, denoted as $B_i = (x_i, y_i, w_i, h_i)$ for $i \in \{1, 2\}$. Additionally, to obtain multiple local representations, several such boxes are dynamically generated, resulting in $B_1^k$ and $B_2^k$, where $k = \{1, \ldots, K\}$ and $K$ is the number of local boxes.

After obtaining the corresponding matching bounding boxes $B_1^k$ and $B_2^k$ during the data augmentation step, we crop the corresponding regions from the feature maps, referred to as the regions of interest (RoIs). Local pooling is then performed on the cropped feature maps using 1×1 RoI Align, resulting in $r_1^k = \mathbf{RoIAlign}(B_1^k, m_1^d)$ and $r_2^k = \mathbf{RoIAlign}(B_2^k, m_2^d)$, which are termed dynamic instances.

Following the MoCoV2 baseline, the representations $r_1^k$ and $r_2^k$ are passed through a projection head to obtain $z_1^k$ and $z_2^k$. In the DI module, a prediction head is additionally applied to the online network, which maps $z_1^k$ to $f_1^k$ in order to predict $z_2^k$ from the target network. This encourages the alignment of local semantic information and improves the spatial representation capability of the network.

The loss of DI module is defined as the mean squared error between the normalized predictions and the normalized target projections, as follows:
\begin{equation}
L_l=\frac{1}{K} \sum_{k=1}^K\left\|\overline{f_1^k}-\overline{z_2^k}\right\|_2^2
\end{equation}
where $\overline{f_1^k}=f_1^k /\left\|f_1^k\right\|_2$ and $ \overline{z_2^k}=z_2^k /\left\|z_2^k\right\|_2$, for normalizing representation vectors.

\subsubsection{Contour Consistency Module}
We found that shallow CNN features are particularly responsive to SAR image contours and provide reliable discriminative cues even in the presence of speckle noise. Therefore, the CC module is designed to incorporate shallow features into the model's attention to SAR land-cover boundaries.

Specifically, by applying $(\mathbf{f_{\theta},f_{\xi}})$ to the augmented views $v_1$ and $v_2$, we obtain the shallow feature maps $m_1^{s}$ and $m_2^s$. Similar to the processing of deep features, global average pooling (GAP) and a projection head are applied to obtain $p_1^s$ and $p_2^s$. Subsequently, the CC module loss is computed, defined as follows:

\begin{equation}
L_s=-\log \frac{\exp \left(p_1^s \cdot p_2^s / \tau\right)}{\exp \left(p_1^s \cdot p_2^s / \tau\right)+\sum_{i=0}^M \exp \left(p_1^s \cdot k_i^s / \tau\right)}
\end{equation}

\subsubsection{Self-Supervised pre-training}
The Dynamic Instance module promotes global context awareness by enforcing local consistency across the image, while the Contour Coherence module guides the model to focus on shapes and geometric structure. The collaboration between them enables the model to attend to the global context of the image while aligning its focus with land-cover objects. Therefore, the overall loss function of the DI3CL framework is defined as:
\begin{equation}
 L = \alpha L_d + (1 - \alpha) L_s + \beta L_l
 \end{equation}
 where $ \alpha$ and $ \beta $ are hyperparameters used to balance the losses. In this study, $\alpha $ is set to 0.8 and $ \beta $ is set to 10 by default to align the magnitudes of different losses.
 
As mentioned above, gradients are propagated only through the online network, while the parameters $\xi$ of the target network are updated using the exponential moving average of the online network parameters $\theta$.

\begin{table*}[ht]
\setcounter{table}{2}
\centering
\caption{Comparison results with random initialization, Imagenet pre-training, and other CL methods on downstream SAR land-cover classification dataset}
\label{results}
\begin{tabular}{c | c c c c c c c c c c| c c c}
\hline
\multirow{2}{*}{Methods} & \multicolumn{10}{c|}{$\mathrm{F1}$ score of each category(\%)}                                                                          & \multirow{2}{*}{OA(\%)} & \multirow{2}{*}{Kappa(\%)} & \multirow{2}{*}{mIoU(\%)} \\
                         & water & wetland & farmland &  forest &  meadow         & road    & village & city  & ground & other &                         &                            &                       \\
\hline
Random Init              & 44.96 & 12.34    & 74.57    & 88.96   & 81.50           & 2.37     & 63.90   & 64.78 & 41.33  & 0.00        & 79.88                   & 73.61                      & 42.30                 \\
ImageNet                 & \textbf{56.23}   & 30.58    & 78.92   & 91.24           & 85.21    & 25.11   & 68.95   & 69.48 & 56.45  & 9.67        & 83.23                   & 78.22                      & 49.49                 \\
\hline
SimCLR                   & 52.65 & 27.35    & 78.49    & 91.05   & 84.88           & 24.34    & 67.42   & 67.97 & 55.19 & 12.18        & 82.84                   & 77.70                      & 48.52                 \\
MoCoV1                   & 53.35 & 24.45    & 78.55    & 91.24   & 85.21           & 20.32    & 68.88   & 68.35 & 55.56  & 9.85        & 83.12                   & 78.06                      & 48.37                 \\
MoCoV2                   & 54.36 & 25.25    & 78.75    & 91.17   & 85.16           & 24.86    & 69.18   & 69.16 & 58.89  & 6.56        & 83.17                   & 78.21                      & 49.02                 \\
BYOL                     & 52.22 & 23.54    & 77.84    & 90.80   & 84.73           & 18.71    & 67.93   & 68.01 & 56.82  & 6.16        & 82.67                   & 77.46                      & 47.69                 \\
SimSiam                  & 52.91 & 25.19    & 78.83    & 91.25   & 86.64           & 24.28   & 69.75   & 69.45 & 54.96  & 7.99        & 83.46                   & 78.45                      & 48.77                 \\
Barlow Twins             & 54.60 & 28.15    & 79.32    & 91.44   & 85.95           & 24.76   & 68.76   & 68.17 & 57.64 & 12.23        & 83.66                   & 78.76                      & 49.45                 \\
PixPro                   & 51.82 & 26.87    & 78.74    & 91.06   & 85.37           & 27.25    & 69.61   & 68.20 & 54.83  & 9.39        & 83.19                   & 78.17                      & 48.79                 \\
Resim                    & 52.69 & 25.70    & 78.75    & 91.07   & 85.34           & 27.81   & 69.58   & 68.96 & 56.07 & 12.16        & 83.14                   & 78.12                      & 49.15                 \\
SSCCL                   & 55.95 & 29.30     &79.57     &91.60    &\textbf{86.69}            & 26.40   & 69.93   & 69.94 & 58.18  & 10.56       & 84.04                  & 79.26                        & 50.18  \\
DINOv2                  & 47.72     & 23.91         & 77.53        & 90.01        & 82.90                       & 17.00       & 66.20       & 65.49     & 46.82      & 5.63           & 81.49                      & 75.83                             & 45.55       \\ 
DI3CL(ours)             & 53.84 &\textbf{32.13}&\textbf{80.19}&\textbf{91.73}& 86.57&\textbf{28.89} & \textbf{71.19}   & \textbf{70.32} & \textbf{59.71} & \textbf{16.28}        & \textbf{84.25}                   & \textbf{79.57}                      & \textbf{51.03}                 \\       
\hline
\end{tabular}
\end{table*}

\begin{figure*}[htbp]
    \centering
    \includegraphics[width=1.00\textwidth]{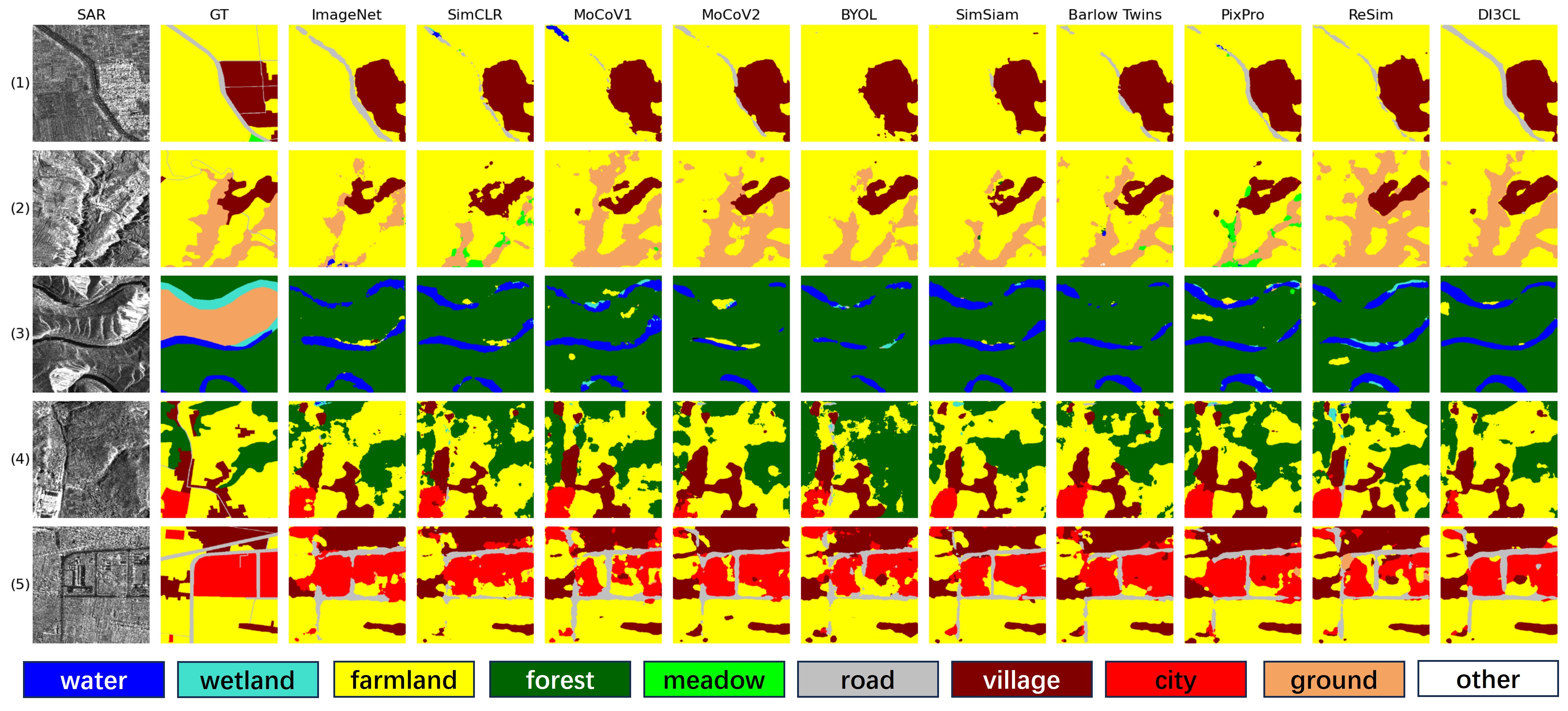} 
    \caption{Visualization results of our method and compared CL methods on downstream SAR land-cover classification dataset.}
    \label{fig:figure_a}
\end{figure*}

\section{EXPERIMENTS}

To evaluate the effectiveness of our proposed improvements to CL-based pre-training methods for SAR land-cover classification tasks, we first conduct a comparative study against several representative CL frameworks, including SimCLR \textcolor{blue}{\cite{simclr}}, MoCo \textcolor{blue}{\cite{moco}}, MoCoV2 \textcolor{blue}{\cite{mocov2}}, BYOL \textcolor{blue}{\cite{byol}}, SimSiam \textcolor{blue}{\cite{simsiam}}, Barlow Twins \textcolor{blue}{\cite{barlowtwins}}, and DINOv2 \textcolor{blue}{\cite{DINOv2}}. We also compare with several local feature enhancement methods such as PixPro \textcolor{blue}{\cite{pixpro}}, Resim \textcolor{blue}{\cite{resim}}, and SSCCL \textcolor{blue}{\cite{ssccl}}. A relatively small-scale pre-training dataset is employed not only to enable fast validation across all methods but also to facilitate rapid hyperparameter tuning during development.

Subsequently, we scale up pre-training using the complete SARSense dataset to obtain the SAR land-cover classification foundation model. This model is then evaluated across three downstream tasks and compared with a wide range of baselines. These include mainstream computer vision models such as UNet \textcolor{blue}{\cite{UNet}}, FCN \textcolor{blue}{\cite{FCN}}, DANet \textcolor{blue}{\cite{DANet}}, PSPNet \textcolor{blue}{\cite{PSPNet}}, DeepLabV3+ \textcolor{blue}{\cite{deeplab}}, and UPerNet \textcolor{blue}{\cite{UPerNet}}. With the exception of UNet, all models employ ResNet-50 as their backbone network and are initialized using weights pre-trained on ImageNet, which is denoted as IMP. To further validate our approach, we compare it with domain-specific methods for remote sensing, such as FarSeg \textcolor{blue}{\cite{FarSeg}}. Moreover, to further validate the superiority of our pre-trained model, we compare our model against several other pre-trained backbones, including SeCo \textcolor{blue}{\cite{SeCo}}, RSP \textcolor{blue}{\cite{RSP}}, and SMLFR \textcolor{blue}{\cite{SMLFR}} based on the ConvNeXt \textcolor{blue}{\cite{ConvNeXt}} architecture. Additionally, to benchmark against state-of-the-art SAR foundation models, we also validate the performance of SAR-JEPA \textcolor{blue}{\cite{SAR-JEPA}} and SARATR-X \textcolor{blue}{\cite{SARATR-X}} on different downstream tasks. These comprehensive evaluations provide strong evidence of the effectiveness and superiority of our proposed approach.

\subsection{Effectiveness of the DI3CL Pre-training Framework}
\subsubsection{Dataset Description}
To ensure a fair and rigorous comparison of pre-training methodologies, we randomly sample a standardized subset of 60,000 image patches from the full SARSense dataset to serve as a shared pre-training resource, whose details are presented in Table \ref{tab:sarsense_fixed}. Importantly, this subset closely matches the distribution of the parent SARSense dataset, thereby preserving the original data's representativeness. This design eliminates dataset-related biases that might otherwise confound performance comparisons across different pre-training methods. This controlled setup ensures that observed differences in model efficacy are attributable to methodological design rather than to differences in training data scale or diversity.

A dedicated downstream evaluation benchmark is crucial for quantifying the transfer learning efficacy of each pre-trained model. To this end, a small-scale land-cover classification dataset is constructed by randomly sampling an additional 10,000 3m-resolution image patches from SARSense, as shown in Table \ref{tab:sarsense_fixed}. Each patch is manually annotated at the pixel level to cover 10 distinct land-cover classes: water, wetland, farmland, forest, meadow, road, village, city, bare ground, and other. The evaluation dataset is split into training, validation, and test sets at a ratio of 3:1:6. The intentionally large test subset enables robust statistical assessment of pre-training performance, providing sufficient sample size to detect subtle differences in model efficacy and to validate the generalizability of learned representations.

\subsubsection{Experiment Setup}
We perform pre-training using 8 Huawei Ascend 910B NPUs, each with 32 GB of memory and implement the proposed DI3CL using the widely used PyTorch framework \textcolor{blue}{\cite{pytorch}}, ensuring efficient computation and optimization.

We select ResNet50 \textcolor{blue}{\cite{resnet}} as the backbone network because CNN architectures remain the mainstream choice for SAR land-cover classification and integrate seamlessly into established workflows. In contrast, while ViT-based \textcolor{blue}{\cite{ViT}} architectures excel at capturing long-range contextual dependencies in generic vision tasks, their performance is susceptible to SAR speckle noise, which requires further exploration in future research. We adopt a specific data augmentation pipeline for contrastive learning, with parameters tailored to the MoCoV2 framework. These settings are based on standard MoCoV2 configurations, optimized for effective contrastive feature learning and tailored to the unique characteristics of SAR images. Data augmentation includes random resized cropping to 448×448 pixels with a scale range of 0.2 to 1.0, random brightness and contrast adjustments (amplitude range: 0.6 to 1.4), Gaussian blur (kernel size: 0.1 to 2.0, probability: 0.5), and random horizontal flipping. Processed images are then converted to tensors and normalized. Model optimization uses stochastic gradient descent with a weight decay of 1e-4 and an initial learning rate of 0.09, following a cosine annealing schedule. Training is performed with multi-NPU parallelism, using a batch size of 32 per NPU (total batch size: 256) over 300 epochs.

We employ the DeepLabV3+ model  \textcolor{blue}{\cite{Deeplabv3}} for downstream SAR land-cover classification, fine-tuning it on two Huawei Ascend 910B NPUs. The model is optimized using stochastic gradient descent with an initial learning rate of 0.03 and a weight decay of 1e-4. We apply the Poly  decay strategy to gradually reduce the learning rate throughout training. The total batch size is set to 64. An early stopping criterion is implemented to halt training if the validation mIoU does not improve for 10 consecutive epochs.

\subsubsection{Experiment Results}
After fine-tuning with a limited amount of labeled data on the downstream land-cover classification dataset, we evaluate the classification performance of the proposed DI3CL pre-training framework. Comparative results are presented in Table \ref{results}. These results show that, when fine-tuned with small labeled data, DI3CL consistently outperforms random initialization, ImageNet pre-training, and other contrastive learning methods. We also benchmarked DI3CL against DINOv2 with a ViT-S backbone, ensuring a comparable parameter count to competing methods. DINOv2 not only lags behind DI3CL but also underperforms several convolution-based approaches, highlighting the inherent limitations of ViT backbones for SAR land-cover classification tasks. Furthermore, DI3CL achieves the highest overall metrics, including OA, Kappa, and mIoU, and consistently delivers the top F1 scores across most categories. This superior performance confirms DI3CL's ability to capture meaningful SAR land-cover features and enhance downstream classification results.

As shown in Fig. \ref{fig:figure_a}, we further perform some visualizations to illustrate the effectiveness of our method. It can be observed that our approach achieves superior classification accuracy, with more complete delineation of elongated objects such as roads and rivers. Additionally, the classification boundaries of other objects are also more precise. These improvements are attributed to the global perception capability introduced by the dynamic instance module and the structural awareness provided by the contour consistency module.

\subsubsection{Ablation Studies}

We conduct ablation studies to systematically assess the impact of DI module box numbers, data augmentation strategies, and core modules on model performance. Results are presented in Tables \ref{ablation_k}, \ref{tab:augmentation_ablation}, and \ref{ablation}. First, we evaluate the performance of DI module with varying numbers of boxes $K$. As shown in Table \ref{ablation_k}, performance steadily improves as $K$ increases from 5 to 15, with the highest OA of 84.25\% and mIoU of 51.03\% at $K=10$, indicating that 10 boxes offer an optimal balance between contextual capture and computational efficiency. Next, we examine the effect of different data augmentation combinations as shown in Table \ref{tab:augmentation_ablation}. Starting from the baseline of Resize and Crop (RC), we find that integrating Color Jitter (CJ) and Gaussian Blur (GB) improves robustness, while adding our DI augmentation produces the most consistent gains across OA, Kappa, and mIoU. The combination of RC, CJ, GB, and DI achieves the best overall performance, confirming that DI effectively mitigates speckle noise and enhances feature generalization.

\begin{table}[t]
\setcounter{table}{3}
\centering
\caption{Ablation study on the number of boxes in the DI module for downstream SAR land-cover classification}
\begin{tabular}{c c c c}
\hline
K & OA(\%) & Kappa(\%) & mIoU(\%) \\
\hline
5 & 84.17 & 79.48 & 50.85 \\
10 & \textbf{84.2}5 & \textbf{79.57} & \textbf{51.03} \\
15 & 83.91 & 79.10 & 50.95 \\
20 & 83.85 & 79.05 & 50.55 \\
\hline
\label{ablation_k}
\end{tabular}
\end{table}

\begin{table}[t]
\setcounter{table}{4}
\centering
\caption{Ablation study on data augmentations for downstream SAR land-cover classification}
\label{tab:augmentation_ablation}
\begin{tabular}{lccc}
\toprule
Augmentations & OA (\%) & Kappa (\%) & mIoU (\%) \\
\midrule
RC                     & 83.30                & 78.16                  & 49.02                  \\
RC+DI                  & 83.59                & 78.72                  & 49.96                  \\
\hline
RC+CJ                  & 83.42                & 78.49                  & 50.07                  \\
RC+CJ+DI               & 83.83                & 79.01                  & 50.83                  \\
\hline
RC+GB                  & 83.49                & 78.58                  & 49.86                  \\
RC+GB+DI               & 84.13                & 79.32                  & 50.56                  \\
\hline
RC+CJ+GB               & 83.96                & 79.19                  & 50.55                  \\
RC+CJ+GB+DI            & 84.25                & 79.57                  & 51.03                  \\
\bottomrule
\end{tabular}

\vspace{0.5em}
{\small
\textbf{Note:} RC: Resize and Crop; CJ: Color Jitter; GB: Gaussian Blur; DI: Dynamic Instances.
}
\end{table}

\begin{table}[t]
\setcounter{table}{5}
\centering
\caption{Ablation study on the effectiveness of the proposed DI and CC modules for downstream SAR land-cover classification}
\begin{tabular}{c c c c}
\hline
Methods & OA(\%) & Kappa(\%) & mIoU(\%) \\
\hline
Baseline(MoCoV2) & 83.17 & 78.21 & 49.02 \\
MoCoV2+DI & 84.03 & 79.31 & 50.73 \\
MoCoV2+DI+CC\_Layer1 & 83.83 & 78.96 & 50.88 \\
MoCoV2+DI+CC\_Layer2 & 83.86 & 79.05 & 50.89 \\
MoCoV2+DI+CC\_Layer3(DI3CL) & \textbf{84.25} & \textbf{79.57} & \textbf{51.03} \\
\hline
\label{ablation}
\end{tabular}
\end{table}

As shown in Table \ref{ablation}, for the core module analysis, suffixes such as “layer$x$” in method names indicate the corresponding residual layer: “layer1” and “layer2” refer to shallow layers, while “layer3” denotes a deep layer. Adding the DI module significantly improves model performance, and incorporating the CC module further enhances effectiveness. Selecting output features from layer3 to construct the CC module yields the best results, serving as the basis for our final model. The DI and CC modules work synergistically, as evidenced by quantitative results. DI alone increases mIoU by 1.71\% over the baseline, and combining CC with DI provides an additional 0.30\% improvement, raising the final mIoU to 51.03\%. This combined gain surpasses the sum of individual module contributions, since the global context capture of DI and geometric sensitivity of CC complement each other to mitigate center bias and enhance feature discrimination in SAR imagery.

\subsection{SAR Land-Cover Classification Foundation Model Pre-training}
\begin{table*}[ht]
\setcounter{table}{6}
\centering
\caption{Comparisons with the SOTA semantic segmentation methods on the test set of downstream SAR land-cover mapping dataset}
\label{all_results}
\begin{tabular}{@{}c|c|cccccccccc|c@{}}
\hline
\multirow{2}{*}{\textbf{Methods}} & \multicolumn{1}{c|}{\multirow{2}{*}{Backbone}} & \multicolumn{10}{c|}{F1 score of each category(\%)}                                                                                                                                                                                                                                                        & \multirow{2}{*}{mIoU(\%)} \\ \cline{3-12}
                                  & \multicolumn{1}{c|}{}                          & \multicolumn{1}{c|}{water} & \multicolumn{1}{c|}{wetland} & \multicolumn{1}{c|}{farmland} & \multicolumn{1}{c|}{forest} & \multicolumn{1}{c|}{meadow} & \multicolumn{1}{c|}{road}  & \multicolumn{1}{c|}{village} & \multicolumn{1}{c|}{city}  & \multicolumn{1}{c|}{ground} & \multicolumn{1}{c|}{other} &                           \\ \hline
\textbf{CV Methods}               & \multicolumn{12}{c}{}                                                                                                                                                                                                                                                                                                                                                                  \\ \hline
UNet                              & \multicolumn{1}{c|}{-}                         & \multicolumn{1}{c|}{65.36} & \multicolumn{1}{c|}{35.19}    & \multicolumn{1}{c|}{81.94}    & \multicolumn{1}{c|}{91.88}  & \multicolumn{1}{c|}{85.90}  & \multicolumn{1}{c|}{30.48} & \multicolumn{1}{c|}{71.99}   & \multicolumn{1}{c|}{70.64} & \multicolumn{1}{c|}{57.84}  & \multicolumn{1}{c|}{2.48 }  & 51.84                    \\ 
FCN                               & \multicolumn{1}{c|}{IMP-ResNet50}              & \multicolumn{1}{c|}{64.69} & \multicolumn{1}{c|}{41.01}    & \multicolumn{1}{c|}{81.53}    & \multicolumn{1}{c|}{91.86}  & \multicolumn{1}{c|}{86.52}  & \multicolumn{1}{c|}{34.29} & \multicolumn{1}{c|}{71.87}   & \multicolumn{1}{c|}{70.93} & \multicolumn{1}{c|}{61.80} & \multicolumn{1}{c|}{26.53}  & 54.09                     \\ 
DANet                             & \multicolumn{1}{c|}{IMP-ResNet50}              & \multicolumn{1}{c|}{65.98} & \multicolumn{1}{c|}{39.85}    & \multicolumn{1}{c|}{81.98}    & \multicolumn{1}{c|}{92.56}  & \multicolumn{1}{c|}{87.61}  & \multicolumn{1}{c|}{34.81} & \multicolumn{1}{c|}{72.09}   & \multicolumn{1}{c|}{71.84} & \multicolumn{1}{c|}{66.28} & \multicolumn{1}{c|}{38.11}  & 55.80                     \\ 
PSPNet                            & \multicolumn{1}{c|}{IMP-ResNet50}              & \multicolumn{1}{c|}{65.46} & \multicolumn{1}{c|}{40.53}    & \multicolumn{1}{c|}{81.57}    & \multicolumn{1}{c|}{92.33}  & \multicolumn{1}{c|}{85.90}  & \multicolumn{1}{c|}{35.48} & \multicolumn{1}{c|}{72.32}   & \multicolumn{1}{c|}{73.56} & \multicolumn{1}{c|}{64.38} & \multicolumn{1}{c|}{23.15}  & 54.59                     \\ 
DeepLabV3+                        & \multicolumn{1}{c|}{IMP-ResNet50}              & \multicolumn{1}{c|}{63.97} & \multicolumn{1}{c|}{43.47}    & \multicolumn{1}{c|}{81.62}    & \multicolumn{1}{c|}{92.45}  & \multicolumn{1}{c|}{87.61}  & \multicolumn{1}{c|}{31.87} & \multicolumn{1}{c|}{71.50}   & \multicolumn{1}{c|}{71.85} & \multicolumn{1}{c|}{65.31} & \multicolumn{1}{c|}{22.84}  & 54.49                     \\ 
UPerNet                           & \multicolumn{1}{c|}{IMP-ResNet50}              & \multicolumn{1}{c|}{64.77} & \multicolumn{1}{c|}{45.90}    & \multicolumn{1}{c|}{81.76}    & \multicolumn{1}{c|}{92.26}  & \multicolumn{1}{c|}{87.64}  & \multicolumn{1}{c|}{38.04} & \multicolumn{1}{c|}{72.13}   & \multicolumn{1}{c|}{71.96} & \multicolumn{1}{c|}{64.99} & \multicolumn{1}{c|}{20.86}  & 55.11                     \\ \hline
\textbf{RS Methods}               & \multicolumn{12}{c}{}                                                                                                                                                                                                                                                                                                                                                                  \\ \hline
FarSeg                            & \multicolumn{1}{c|}{IMP-ResNet50}              & \multicolumn{1}{c|}{62.99} & \multicolumn{1}{c|}{40.16}    & \multicolumn{1}{c|}{81.10}    & \multicolumn{1}{c|}{92.19}  & \multicolumn{1}{c|}{87.36}  & \multicolumn{1}{c|}{35.66} & \multicolumn{1}{c|}{71.81}   & \multicolumn{1}{c|}{70.42} & \multicolumn{1}{c|}{64.23} & \multicolumn{1}{c|}{17.46}  & 53.72                     \\ 
UPerNet                           & \multicolumn{1}{c|}{SeCo-ResNet50}             & \multicolumn{1}{c|}{64.86} & \multicolumn{1}{c|}{43.81}    & \multicolumn{1}{c|}{82.02}    & \multicolumn{1}{c|}{92.47}  & \multicolumn{1}{c|}{87.92}  & \multicolumn{1}{c|}{36.21} & \multicolumn{1}{c|}{71.81}   & \multicolumn{1}{c|}{71.93} & \multicolumn{1}{c|}{66.09} & \multicolumn{1}{c|}{24.86}  & 55.24                    \\ 
UPerNet                           & \multicolumn{1}{c|}{RSP-ResNet50}              & \multicolumn{1}{c|}{66.62} & \multicolumn{1}{c|}{42.84}    & \multicolumn{1}{c|}{81.82}    & \multicolumn{1}{c|}{92.44}  & \multicolumn{1}{c|}{87.17}  & \multicolumn{1}{c|}{35.07} & \multicolumn{1}{c|}{72.79}   & \multicolumn{1}{c|}{72.18} & \multicolumn{1}{c|}{65.05} & \multicolumn{1}{c|}{20.68}  & 54.93                     \\ 
UPerNet                           & \multicolumn{1}{c|}{SMLFR+ConvNeXt-B}          & \multicolumn{1}{c|}{66.55} & \multicolumn{1}{c|}{45.32}    & \multicolumn{1}{c|}{82.26}    & \multicolumn{1}{c|}{92.43}  & \multicolumn{1}{c|}{87.09}  & \multicolumn{1}{c|}{40.89} & \multicolumn{1}{c|}{73.06}   & \multicolumn{1}{c|}{73.03} & \multicolumn{1}{c|}{67.31} & \multicolumn{1}{c|}{37.69}  & 56.95                     \\
UPerNet                           & \multicolumn{1}{c|}{SAR-JEPA+ViT-B}            & \multicolumn{1}{c|}{61.72}     & \multicolumn{1}{c|}{33.72}        & \multicolumn{1}{c|}{78.88}        & \multicolumn{1}{c|}{90.60}      & \multicolumn{1}{c|}{85.19}      & \multicolumn{1}{c|}{21.79}     & \multicolumn{1}{c|}{68.34}        & \multicolumn{1}{c|}{68.73}     & \multicolumn{1}{c|}{58.71}    & \multicolumn{1}{c|}{22.40}      & 50.67                          \\
UPerNet                           & \multicolumn{1}{c|}{SARATR-X+HiViT-B}            & \multicolumn{1}{c|}{63.12}     & \multicolumn{1}{c|}{45.12}        & \multicolumn{1}{c|}{81.85}        & \multicolumn{1}{c|}{92.18}      & \multicolumn{1}{c|}{87.45}      & \multicolumn{1}{c|}{34.29}     & \multicolumn{1}{c|}{72.46}        & \multicolumn{1}{c|}{70.47}     & \multicolumn{1}{c|}{65.69}    & \multicolumn{1}{c|}{31.51}      & 55.18                          \\ 
\hline
\textbf{Ours}                     & \multicolumn{12}{c}{}                                                                                                                                                                                                                                                                                                                                                                  \\ \hline
UPerNet                           & \multicolumn{1}{c|}{DI3CL-ResNet50}            & \multicolumn{1}{c|}{69.54} & \multicolumn{1}{c|}{50.88}    & \multicolumn{1}{c|}{83.38}    & \multicolumn{1}{c|}{\textbf{93.39}}  & \multicolumn{1}{c|}{\textbf{88.76}}  & \multicolumn{1}{c|}{40.96} & \multicolumn{1}{c|}{73.74}   & \multicolumn{1}{c|}{73.55} & \multicolumn{1}{c|}{\textbf{70.28}} & \multicolumn{1}{c|}{\textbf{35.83}}  & 58.59                     \\ 
UPerNet                           & \multicolumn{1}{c|}{DI3CL-ResNet101}           & \multicolumn{1}{c|}{\textbf{70.18}} & \multicolumn{1}{c|}{\textbf{51.32}}    & \multicolumn{1}{c|}{\textbf{83.47}}    & \multicolumn{1}{c|}{93.33}  & \multicolumn{1}{c|}{88.68}  & \multicolumn{1}{c|}{\textbf{43.99}} & \multicolumn{1}{c|}{\textbf{73.96}}   & \multicolumn{1}{c|}{\textbf{73.84}} & \multicolumn{1}{c|}{69.51} & \multicolumn{1}{c|}{32.52}  & \textbf{58.67}                     \\ \hline

\end{tabular}
\end{table*}

\begin{figure*}[htbp]
    \centering
    \includegraphics[width=1.00\textwidth]{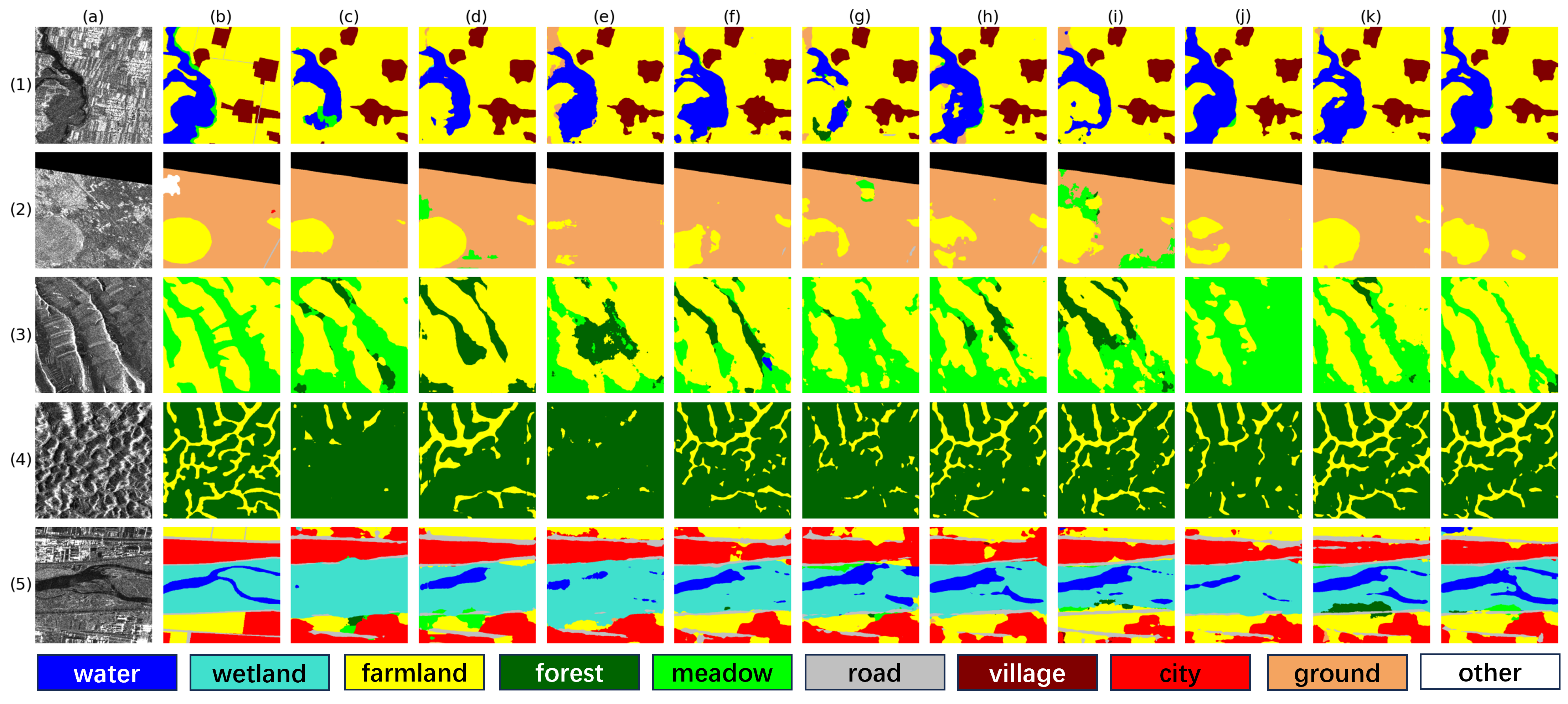} 
    \caption{Visualization results of our method and compared methods on SAR land-cover mapping dataset. (a) SAR. (b) GT. (c) DANet. (d) PSPNet. (e) DeepLabV3+. (f) UPerNet. (g) FarSeg. (h) UPerNet-RSP. (i) UPerNet-SMLFR. (j) UPerNet-SARATR-X. (k) UPerNet-DI3CL-R50. (l) UPerNet-DI3CL-R101.}
    \label{figure_all_element}
\end{figure*}
We use the entire SARSense dataset and adopt the DI3CL pre-training framework to train our SAR land-cover classification foundation model. ResNet50 and ResNet101 are chosen as the backbone architectures of our foundation model due to their strong feature representation capabilities and widespread application in various computer vision tasks. The parameter settings described in Section IV-A are followed. To accelerate the training process, a multi-node parallel training strategy is applied, where pre-training is conducted on four nodes, each equipped with 8 Huawei Ascend 910B NPUs, resulting in a total batch size of 1024. Pre-training is conducted for 300 epochs, and the weights from the epoch with the lowest training loss are used to construct the final foundation model. Finally, task-specific classification heads can be attached to the backbone network, enabling adaptation to various downstream classification tasks. In this study, we conduct three SAR land-cover classification experiments, including SAR land-cover mapping, SAR water detection, and SAR road extraction. The results demonstrate the effectiveness of our SAR foundation model.

\subsection{SAR Land-Cover Mapping}

\subsubsection{Dataset Description}
A total of 33,108 3m-resolution images are selected from the SARSense dataset for meticulous manual annotation, whose details are presented in Table \ref{tab:sarsense_fixed}. This carefully curated dataset is designed to rigorously assess the performance of our SAR foundation model on practical land-cover mapping tasks. To ensure robust model training and evaluation, the data is partitioned into training, validation, and test sets at 8:1:1, providing ample samples for multi-class land-cover classification: 26,465 for training, 3,279 for validation, and 3,364 for testing.

\subsubsection{Experimental Setup}
To ensure consistency across experiments, UPerNet is adopted as the semantic segmentation head. Additionally, to benchmark against existing SAR foundation models, pre-trained weights of SAR-JEPA and SARATR-X are loaded and fine-tuned across diverse downstream tasks to evaluate their performance. All experiments are conducted on 4 Huawei Ascend 910B NPUs, with models fine-tuned using the stochastic gradient descent optimizer, an initial learning rate of 0.01, and a weight decay of 1e-4. Similar to the settings in Section IV-A, the Poly learning rate decay strategy is applied to progressively reduce the learning rate during training, with a batch size of 32 per device.

For evaluation, the F1-score is calculated for each class to assess class-wise classification accuracy, while the mIoU is used as the overall performance metric. An early stopping strategy is also employed, where training is terminated if the mIoU on the validation set does not improve for 10 consecutive epochs.

\subsubsection{Experimental Results}
After fine-tuning the models on the training and validation sets, we evaluate their performance on the test set. The results, shown in Table \ref{all_results}, offer a comprehensive comparison of various methods. Notably, foundation models designed specifically for SAR object recognition tasks underperform, struggling to extract fine land-cover details. In contrast, our proposed model consistently achieves the highest F1 scores across all land-cover categories. For the challenging wetland class, the DI3CL-ResNet101 model attains an F1 score of 51.32\%, representing a significant improvement over existing approaches. This performance boost is primarily attributed to the strong contour-awareness capability integrated into our model's design. Additionally, our model achieves the highest mIoU, further demonstrating its robustness and effectiveness in complex multi-class land-cover mapping scenarios. 

In addition to quantitative evaluation, we also provide qualitative visualizations of the prediction results, as illustrated in Fig. \ref{figure_all_element}. It is evident that our proposed DI3CL model achieves higher classification accuracy compared to other methods, with fewer misclassifications and improved consistency and precision along object boundaries. These results powerfully demonstrate the superior suitability of our model for SAR land-cover classification tasks.

\subsection{SAR Water Detection}
\begin{table}[ht]
\centering
\setcounter{table}{7}
\tiny
\caption{Comparisons with the SOTA semantic segmentation methods on the test set of SAR water detection dataset}
\label{water_results}
\begin{tabular}{@{}c|c|c|c|c|c}
\hline
\textbf{Methods}                  & \multicolumn{1}{c|}{Backbone}                  & Precision(\%)      & Recall(\%)    & F1 Score(\%)     & IoU(\%)\\
\hline
\textbf{CV Methods}               & \multicolumn{5}{c}{} \\ \hline
FCN                               & \multicolumn{1}{c|}{IMP-ResNet50}              & 76.60              & 67.84         & 73.09            & 57.59                     \\ 
DANet                             & \multicolumn{1}{c|}{IMP-ResNet50}              & 74.23              & 70.56         & 72.35            & 56.68                     \\ 
PSPNet                            & \multicolumn{1}{c|}{IMP-ResNet50}              & 78.85              & 71.80         & 75.16            & 60.20                     \\ 
DeepLabV3+                        & \multicolumn{1}{c|}{IMP-ResNet50}              & 70.67              & 74.47 & 72.52           & 56.89                     \\ 
UPerNet                           & \multicolumn{1}{c|}{IMP-ResNet50}              & 72.38              & 72.20         & 72.29            & 56.60                   \\ \hline
\textbf{RS Methods}               & \multicolumn{5}{c}{}                                                                                                                                          \\ \hline
FarSeg                            & \multicolumn{1}{c|}{IMP-ResNet50}              & 75.89              & 71.36         & 73.55            & 58.17                   \\ 
UPerNet                           & \multicolumn{1}{c|}{SeCo-ResNet50}             & 73.14              & 68.86         & 70.94            & 54.96                   \\ 
UPerNet                           & \multicolumn{1}{c|}{RSP-ResNet50}              & 71.93              & 72.15         & 72.04            & 56.30                   \\ 
UPerNet                           & \multicolumn{1}{c|}{SMLFR+ConvNeXt-B}          & 75.18              & 69.18         & 72.05            & 56.37                  \\ 
UPerNet                           & \multicolumn{1}{c|}{SAR-JEPA+ViT-B}            & 75.99              & 71.36         & 73.60            & 58.23                   \\ 
UPerNet                           & \multicolumn{1}{c|}{SARATR-X+HiViT-B}          & 72.23              & \textbf{75.23}         & 73.70            & 58.35                       \\ \hline
\textbf{Ours}                     & \multicolumn{5}{c}{}                                                                                                                                      
\\ \hline
UPerNet                           & \multicolumn{1}{c|}{DI3CL-ResNet50}            & \textbf{80.09}              & 71.49        & \textbf{75.55}            & \textbf{60.70}                     \\ 
UPerNet                           & \multicolumn{1}{c|}{DI3CL-ResNet101}           & 79.48     & 71.37         & 75.21            & 60.26                     \\ \hline

\end{tabular}
\end{table}

\begin{figure*}[htbp]
    \centering
    \includegraphics[width=1.00\textwidth]{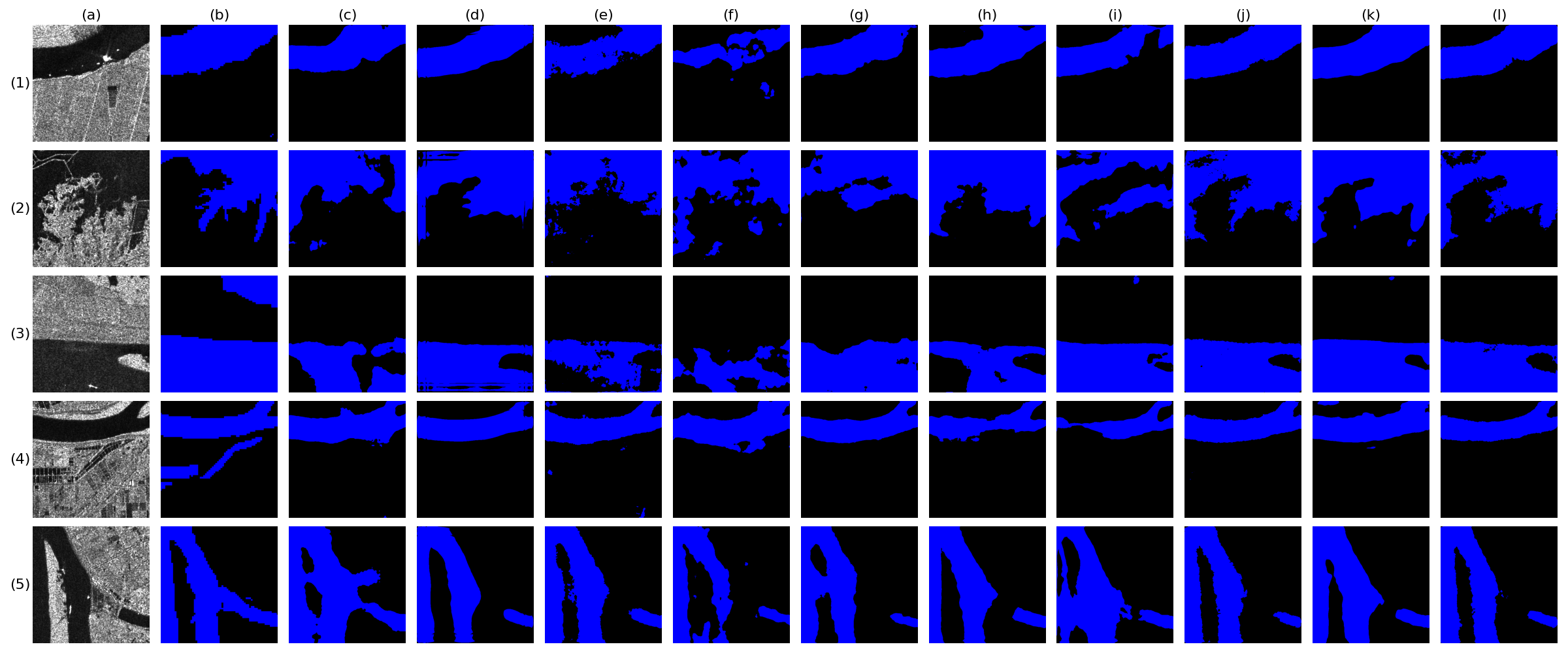} 
    \caption{Visualization results of our method and compared methods on water detection dataset.(a) SAR. (b) GT. (c) DANet. (d) PSPNet. (e) DeepLabV3+. (f) UPerNet. (g) FarSeg. (h) UPerNet-RSP. (i) UPerNet-SMLFR. (j) UPerNet-SARATR-X. (k) UPerNet-DI3CL-R50. (l) UPerNet-DI3CL-R101}
    \label{figure_water}
\end{figure*}

\subsubsection{Dataset Description}

To validate the water detection capability of SAR foundation models, a dedicated dataset is constructed, independently curated, and entirely disentangled from the SARSense dataset, to rigorously assess the generalization performance of our foundation model across diverse downstream tasks. As shown in Table \ref{tab:sarsense_fixed}, the dataset is acquired exclusively by the GaoFen-3 satellite, with a spatial resolution of 5 meters and HH polarization. For experimentation, the data are cropped into patches of 512×512 pixels, including training, validation, and test sets.  The training and validation sets are constructed from 11 SAR image scenes collected from regions such as Shaanxi, Jiangxi, and Anhui, and the image patches are randomly shuffled and split into training and validation sets in a 9:1 ratio, resulting in 9,819 training samples and 1,091 validation samples. The test set is independently constructed from 4 image scenes collected from Hunan, Hubei, and Henan provinces, yielding a total of 6,593 test samples after cropping, with the training and validation sets remaining completely disjoint from the test set.

\subsubsection{Experimental Setup}

Similarly, in this experimental setting, we employ UPerNet as the decoder head for the water detection task. Considering the limited scale of the dataset and the goal of minimizing computational overhead, we conduct all experiments using two Huawei Ascend 910B NPUs.

The training setup largely follows that of the SAR land-cover mapping task, with the primary difference being the choice of the initial learning rate, which is set to 0.03 in this case, to better suit the data characteristics and convergence behavior.

For the evaluation phase, we adopt multiple standard metrics to comprehensively assess the performance of water detection. These include Precision, Recall, F1 Score, and IoU, which together provide a detailed understanding of the model’s accuracy and robustness in identifying water regions.

\subsubsection{Experimental Results}
The test set results are summarized in Table \ref{water_results}. Notably, models such as UPerNet and DeepLabV3+, which perform well in SAR land-cover mapping, show comparatively weaker performance on this specific task, while SAR-JEPA and SARATR-X achieve only moderate results. This decline is mainly due to overfitting on the training data and substantial distributional differences between the training and test sets, which hinder the models' generalization ability. This finding is further supported by the initialization strategy for the UPerNet segmentation head, where weights pre-trained on remote sensing datasets produce poorer results than those initialized with ImageNet pre-trained weights.

In contrast, our proposed model achieves a compelling balance between strong fitting capacity and robust generalization ability. Except for recall, our model attains the highest performance across all evaluation metrics, indicating its superior detection capability for the water class. Interestingly, the results obtained using ResNet101 as the backbone are slightly lower than those with ResNet50, which can similarly be explained by the overfitting and generalization trade-off discussed above.

We also select several samples from the water detection task for qualitative visualization. As shown in Fig. \ref{figure_water}, our model produces more complete detection results that more closely align with the actual water regions. Overall, the experimental findings demonstrate that our model, without relying on any domain adaptation or domain generalization techniques, can still achieve excellent generalization performance. This not only validates the model’s effectiveness in accurately detecting water but also highlights its inherent robustness when faced with domain shifts between training and testing data.

\begin{figure*}[htbp]
    \centering
    \includegraphics[width=1.05\textwidth]{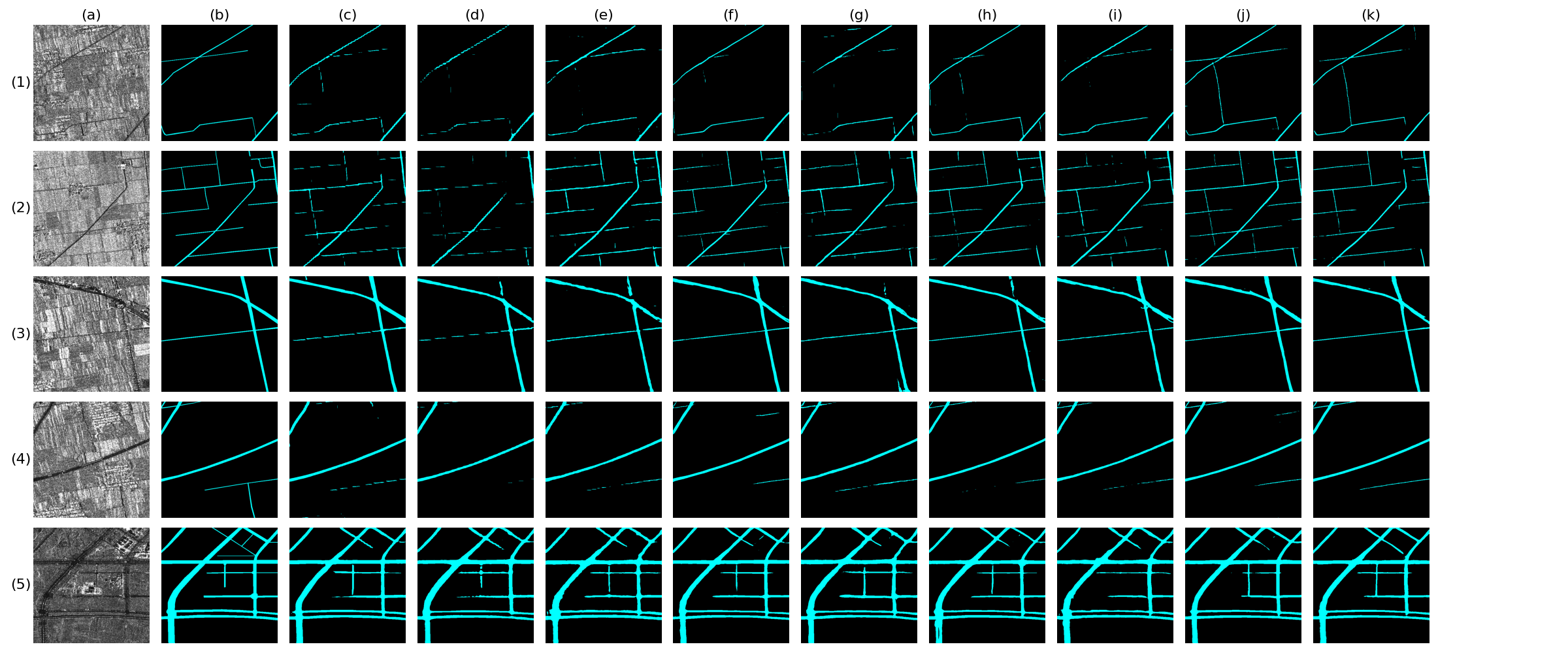} 
    \caption{Visualization results of our method and compared methods on road extraction dataset.(a) SAR. (b) GT. (c) DANet. (d) PSPNet. (e) DeepLabV3+. (f) UPerNet. (g) FarSeg. (h) UPerNet-RSP. (i) UPerNet-SARATR-X. (j) UPerNet-DI3CL-R50. (k) UPerNet-DI3CL-R101}
    \label{figure_road}
\end{figure*}

\subsection{SAR Road Extraction}

\subsubsection{Dataset Description}

To evaluate SAR foundation models for road extraction, a dedicated dataset is created and curated independently of SARSense. As shown in Table \ref{tab:sarsense_fixed}, the dataset is derived exclusively from GaoFen-3 satellite imagery with a spatial resolution of 3 meters and HH single polarization, sourced from three SAR scenes covering representative areas in the Guanzhong region of Shaanxi Province, which is characterized by flat terrain, high urbanization, and a well-developed road network. Each large scene is cropped into 512×512 pixel patches, followed by data cleaning to remove images with low pixel proportion of road or significant geometric distortion. After filtering, 1,107 patches are retained and manually annotated with high precision. The dataset is randomly split into training, validation, and test sets at 8:1:1, yielding 886 training, 110 validation, and 111 test samples. Given its relatively small size that elevates overfitting risk, data augmentation techniques are employed for training samples, including 90°, 180°, 270° rotations and horizontal/vertical flipping, expanding the training set to sixfold its original size, with the final dataset comprising 5,316 training, 110 validation, and 111 test samples.

\subsubsection{Experimental Setup}
The experimental configuration for this task is broadly consistent with that of the SAR water detection experiments. The primary differences lie in two aspects: first, the initial learning rate is set to 0.08 in this task to better accommodate the training dynamics; second, due to the inherent class imbalance in the dataset, we additionally incorporate the Dice loss \textcolor{blue}{\cite{Dice}} function alongside the standard loss to mitigate the adverse effects of uneven class distributions.

\begin{table}[t]
\setcounter{table}{8}
\centering
\tiny
\caption{Comparisons with the SOTA semantic segmentation methods on the test set of SAR road extraction dataset}
\label{road_results}
\begin{tabular}{@{}c|c|c|c|c|c}
\hline
\textbf{Methods}                  & \multicolumn{1}{c|}{Backbone}                  & Precision(\%)      & Recall(\%)    & F1 Score(\%)     & IoU(\%)\\
\hline
\textbf{CV Methods}               & \multicolumn{5}{c}{} \\ \hline
UNet                              & \multicolumn{1}{c|}{-}                         & 74.14              & 66.16         & 69.92            & 53.75                     \\ 
FCN                               & \multicolumn{1}{c|}{IMP-ResNet50}              & 73.27              & 70.07         & 71.63            & 55.80                    \\ 
DANet                             & \multicolumn{1}{c|}{IMP-ResNet50}              & 70.81              & 72.74         & 71.76            & 55.96                     \\ 
PSPNet                            & \multicolumn{1}{c|}{IMP-ResNet50}              & 75.95              & 64.93         & 70.01            & 53.86                  \\ 
DeepLabV3+                        & \multicolumn{1}{c|}{IMP-ResNet50}              & 70.27              & 72.66         & 71.44            & 55.57                     \\ 
UPerNet                           & \multicolumn{1}{c|}{IMP-ResNet50}              & 75.78              & 70.54         & 73.07            & 57.56                   \\ \hline
\textbf{RS Methods}               & \multicolumn{5}{c}{}                                                                                                                                                                                                                                                                                                                                    \\ \hline
FarSeg                            & \multicolumn{1}{c|}{IMP-ResNet50}              & 70.29              & 69.06         & 69.67            & 53.46                    \\ 
UPerNet                           & \multicolumn{1}{c|}{SeCo-ResNet50}             & 74.82              & 68.32         & 71.42            & 55.55                    \\ 
UPerNet                           & \multicolumn{1}{c|}{RSP-ResNet50}              & 77.28              & 69.23         & 73.03            & 57.52                    \\ 
UPerNet                           & \multicolumn{1}{c|}{SAR-JEPA+ViT-B}            & 70.34              & 60.16         & 64.85            & 48.00                      \\ 
UPerNet                           & \multicolumn{1}{c|}{SARATR-X+HiViT-B}          & 71.78                  & 67.45             & 69.55                & 53.31                       \\\hline
\textbf{Ours}                     & \multicolumn{5}{c}{}                                                                                                                                                                                                                                                                                                                                    
\\ \hline
UPerNet                           & \multicolumn{1}{c|}{DI3CL-ResNet50}            & 74.83              & \textbf{72.90}         & 73.85            & 58.54                     \\ 
UPerNet                           & \multicolumn{1}{c|}{DI3CL-ResNet101}           & \textbf{77.76}     & 71.70        & \textbf{74.60}            & \textbf{59.50}                     \\ \hline

\end{tabular}
\end{table}

\subsubsection{Experimental Results}

The evaluation results of the SAR road extraction task are summarized in Table \ref{road_results}. Notably, SAR-JEPA and SARATR-X perform poorly on this structure and detail-demanding road extraction task. As shown, our model achieves the highest scores across all metrics except precision. Compared to the second-best model, UPerNet with ImageNet initialization, our method outperforms it by 0.98\% in IoU and a 0.78\% gain in F1 score. These results underscore the effectiveness of our SAR-tailored model and confirm its superiority in accurately segmenting road structures from SAR imagery.

Fig.\ref{figure_road} illustrates qualitative comparisons of road extraction results from different models on the SAR image test set. Several representative examples from diverse scenes are selected to visually showcase the performance differences. It can be observed that our model demonstrates the strongest capability in road extraction from SAR imagery. The extraction results for narrow roads are highly continuous with minimal noise. In urban areas, wide roads are also effectively identified, with well-balanced road boundaries and clearly segmented intersections. Roads of various scales are successfully extracted, indicating the adaptability and effectiveness of the proposed foundation model for the road extraction task.

\subsection{Efficiency Comparison}
We further conduct an efficiency comparison across representative SAR foundation models, as summarized in Table \ref{tab:model_comparison}. Our Res50 and Res101 variants achieve a favorable balance between computational cost and performance. Notably, the Res101 model, with 44M parameters, is substantially more parameter-efficient than its ViT-based counterparts, SAR-JEPA (85M) and SARATR-X (66M). For pre-training efficiency, our Res50 model completes 300 epochs in just 40 hours on Ascend910B hardware. Our models are optimized for Ascend910B infrastructure, with VRAM usage of 18.88G for Res50 and 30.31G for Res101, both comfortably within hardware limits. This optimization avoids the memory bottlenecks frequently encountered by ViT-based models. For inference FPS, our Res50 model delivers substantial inference speed gains, achieving 97.37 FPS, which is 75\% faster than SAR-JEPA (ViT-B) and 25\% faster than SARATR-X (HiViT-B). Even with increased capacity, our Res101 variant maintains a strong 84.78 FPS, outperforming both ViT-based models. These results confirm that our approach delivers robust performance without sacrificing computational feasibility, making it well-suited for large-scale SAR land-cover classification tasks.

\begin{table*}[t]
    \centering
    \caption{Efficiency Comparison of SAR Foundation Models in Pre-training and Inference}
    \label{tab:model_comparison}
    \begin{tabular}{lcccccc} 
        \toprule
        \textbf{Models} & \textbf{Params (M)} & \textbf{Pre-training time (h)} & \textbf{Pre-training epochs} & \textbf{Hardware} & \textbf{VRAM} & \textbf{Inference FPS (5090)} \\
        \midrule
        MoCoV2 (Res50)   & 24 & 53 & 200 & 8 $\times$ V100 & $<$ 32G & 97.37 \\
        SAR-JEPA (ViT-B) & 85 & - & 200 & 4 $\times$ V100 & $<$ 32G & 55.65 \\
        SARATR-X (HiViT-B) & 66 & - & 200 & 8 $\times$ RTX 3090 & $<$ 24G & 77.83 \\
        \midrule
        Ours (Res50) & 24 & 40 & 300 & 32 $\times$ Ascend910B  & 18.88G & 97.37 \\
        Ours (Res101)& 44 & 50 & 300 & 32 $\times$ Ascend910B & 30.31G & 84.78 \\
        \bottomrule
    \end{tabular}
\end{table*}

\section{DISCUSSION}
A detailed analysis of the segmentation results reveals specific performance bottlenecks in existing models when handling small-scale water and complex linear road networks. For small water, all models occasionally exhibit under-segmentation or structural discontinuity. Small inland ponds or narrow river tributaries are susceptible to misclassification as adjacent wetlands or are omitted entirely. This limitation is primarily due to the weak, unstable backscatter signals characteristic of small water surfaces, which are frequently obscured by the inherent speckle noise in SAR imagery. From a methodological perspective, existing methods prioritize prominent geometric boundaries in images and use ROIs of varying sizes to maintain instance consistency. However, for very small targets with limited spatial extent, these targets typically occupy only a tiny fraction of the generated patches. As a result, the instance consistency constraint may be dominated by the surrounding background, inadvertently diluting the subtle scattering distinctions of the small water. Furthermore, as foundation models centered on backbone representation learning, the difficulty in recovering these pixel-level details may be partially due to the limitations of standard segmentation decoders in preserving high-frequency spatial information during upsampling.  For complex linear roads, existing models tend to produce fragmented segments rather than continuous networks. This remains a persistent bottleneck in SAR interpretation due to the similarity in backscatter signatures between narrow roads and surrounding bare soil or low-vegetation areas. The image-level instance discrimination task derived from existing methods focuses primarily on broad semantic differentiation rather than fine-grained topological connectivity. Consequently, the learned representations may prioritize category-level robustness over the precise topological continuity required for road extraction.

\section{CONCLUSION}
In this study, a novel CL-based pre-training framework, termed DI3CL, is proposed for SAR land-cover classification. This framework integrates two key modules: the Dynamic Instance (DI) module, which promotes global contextual awareness by enforcing consistency in local representations, and the Contour Consistency (CC) module, which guides the model to attend to geometric structures inherent in SAR imagery. Through adequate experiments, the effectiveness of DI3CL has been validated, demonstrating superior performance over existing CL methods in SAR land-cover classification tasks.

A large-scale dataset of 460,532 SAR images is constructed to pre-train a SAR land-cover foundation model utilizing the DI3CL framework. The foundation model is validated across SAR land-cover mapping, water detection, and road extraction, demonstrating its effectiveness and broad applicability. 

However, the model still has some limitations, particularly in detecting small water and maintaining road continuity. Looking ahead, our immediate focus will be on addressing these limitations. After these improvements, the pre-trained SAR foundation model will be further deployed in practical applications, such as crop monitoring, flood detection, and road network extraction, to more fully demonstrate its practical value. To further improve recall and accuracy, we plan to introduce optical images to exploit the detailed color information of small water and the clear topological features of roads present in optical data. By leveraging these complementary characteristics, this approach aims to enhance semantic understanding, improve the model's discrimination, and address current limitations.

\ifCLASSOPTIONcaptionsoff
  \newpage
\fi


\begin{thebibliography}{50}

\bibitem{all_day1}
Tian Z, Wang W, Zhou K, et al. Weighted pseudo-labels and bounding boxes for semisupervised SAR target detection[J]. IEEE Journal of Selected Topics in Applied Earth Observations and Remote Sensing, 2024, 17: 5193-5203.

\bibitem{all_day2}
Deng J, Wang W, Zhang H, et al. PolSAR Ship Detection Based on Superpixel-Level Contrast Enhancement[J]. IEEE Geoscience and Remote Sensing Letters, 2024.

\bibitem{civil}
Liu Y, Qian J, Yue H. Combined Sentinel-1A with Sentinel-2A to estimate soil moisture in farmland[J]. IEEE Journal of Selected Topics in Applied Earth Observations and Remote Sensing, 2020, 14: 1292-1310.


\bibitem{military}
Song Y, Li J, Gao P, et al. Two-stage cross-modality transfer learning method for military-civilian SAR ship recognition[J]. IEEE Geoscience and Remote Sensing Letters, 2022, 19: 1-5.

\bibitem{monitoring}
X.-M. Li, Y. Sun, and Q. Zhang, “Extraction of sea ice cover by Sentinel-
1 SAR based on support vector machine with unsupervised generation
of training data,” IEEE Trans. Geosci. Remote Sens., vol. 59, no. 4,
pp. 3040–3053, Apr. 2021.

\bibitem{monitoring2}
Wang K, Ren Z, Hou B, et al. BSG-WSL: BackScatter-guided weakly supervised learning for water mapping in SAR images[J]. International Journal of Applied Earth Observation and Geoinformation, 2025, 136: 104385.
\bibitem{monitoring3}
Wang K, Ren Z, Hou B, et al. Water-Matching CAM: A Novel Class Activation Map for Weakly-Supervised Semantic Segmentation of Water in SAR Images[J]. IEEE Journal of Selected Topics in Applied Earth Observations and Remote Sensing, 2024.

\bibitem{agriculture}
Pang J, Zhang R, Yu B, et al. Pixel-level rice planting information monitoring in Fujin City based on time-series SAR imagery[J]. International Journal of Applied Earth Observation and Geoinformation, 2021, 104: 102551.

\bibitem{agriculture2}
Hashemi M G Z, Jalilvand E, Alemohammad H, et al. Review of synthetic aperture radar with deep learning in agricultural applications[J]. ISPRS Journal of Photogrammetry and Remote Sensing, 2024, 218: 20-49.



\bibitem{development}
Manzoni M, Monti-Guarnieri A, Molinari M E. Joint exploitation of spaceborne SAR images and GIS techniques for urban coherent change detection[J]. Remote Sensing of Environment, 2021, 253: 112152.

\bibitem{development2}
Recla M, Schmitt M. The SAR2Height framework for urban height map reconstruction from single SAR intensity images[J]. ISPRS Journal of Photogrammetry and Remote Sensing, 2024, 211: 104-120.

\bibitem{categories}
Ren B, Zhao Y, Hou B, et al. A mutual information-based self-supervised learning model for PolSAR land cover classification[J]. IEEE Transactions on Geoscience and Remote Sensing, 2021, 59(11): 9224-9237.

\bibitem{HA}
S. R. Cloude and E. Pottier, “An entropy based classification scheme for land applications of polarimetric SAR,” IEEE Trans. Geosci. Remote Sens., vol. 35, no. 1, pp. 68–78, Jan. 1997.


\bibitem{freeman}
A. Freeman and S. L. Durden, “A three-component scattering model for polarimetric SAR data,” IEEE Trans. Geosci. Remote Sens., vol. 36, no. 3, pp. 963–973, May 1998.

\bibitem{pauli}
E. Pottier, “Dr. J. R. Huynen’s main contributions in the development of polarimetric radar techniques and how the ’radar targets phenomenological Concept’ becomes a theory,” Proc. SPIE, vol. 1748, pp. 72–85, Feb. 1993.


\bibitem{statistical method1}
Lee J S, Schuler D L, Lang R H, et al. K-distribution for multi-look processed polarimetric SAR imagery[C]//Proceedings of IGARSS'94-1994 IEEE International Geoscience and Remote Sensing Symposium. IEEE, 1994, 4: 2179-2181.

\bibitem{statistical method2}
Freitas C C, Frery A C, Correia A H. The polarimetric $\mathcal{G}$ distribution for SAR data analysis[J]. Environmetrics: The official journal of the International Environmetrics Society, 2005, 16(1): 13-31.

\bibitem{statistical method3}
Song W, Li M, Zhang P, et al. The WG$\Gamma$ distribution for multilook polarimetric SAR data and its application[J]. IEEE Geoscience and Remote Sensing Letters, 2015, 12(10): 2056-2060.


\bibitem{drawbacks}
Huang Z, Yao X, Liu Y, et al. Physically explainable CNN for SAR image classification[J]. ISPRS Journal of Photogrammetry and Remote Sensing, 2022, 190: 25-37.



\bibitem{FCN}
J. Long, E. Shelhamer, and T. Darrell, “Fully convolutional networks for semantic segmentation,” in Proc. IEEE Conf. Comput. Vis. Pattern Recognit., Jun. 2015, pp. 3431–3440.

\bibitem{UNet}
O. Ronneberger, P. Fischer, and T. Brox, “U-Net: Convolutional networks for biomedical image segmentation,” in Proc. Int. Conf. Med. Image Comput. Comput.-Assist. Intervent. Cham, Switzerland: Springer, 2015, pp. 234–241.

\bibitem{Deeplabv1}
L.-C. Chen, G. Papandreou, I. Kokkinos, K. Murphy, and A. L. Yuille, “DeepLab: Semantic image segmentation with deep convolutional nets, atrous convolution, and fully connected CRFs,” IEEE Trans. Pattern Anal. Mach. Intell., vol. 40, no. 4, pp. 834–848, Apr. 2018.

\bibitem{Deeplabv2}
L.-C. Chen, G. Papandreou, F. Schroff, and H. Adam, “Rethinking atrous convolution for semantic image segmentation,” 2017, arXiv:1706.05587.

\bibitem{Deeplabv3}
L.-C. Chen, Y. Zhu, G. Papandreou, F. Schroff, and H. Adam, “Encoder-decoder with atrous separable convolution for semantic image segmentation,” in Proc. Eur. Conf. Comput. Vis. (ECCV), Jun. 2018, pp. 801–818.


\bibitem{CWNN}
Y. Duan, F. Liu, L. Jiao, P. Zhao, and L. Zhang, “SAR image segmentation based on convolutional-wavelet neural network and Markov random field,” Pattern Recognit., vol. 64, pp. 255–267, Apr. 2017.


\bibitem{SAR2}
F. Mohammadimanesh, B. Salehi, M. Mahdianpari, E. Gill, and M. Molinier, “A new fully convolutional neural network for semantic segmentation of polarimetric SAR imagery in complex land cover ecosystem,” ISPRS J. Photogramm. Remote Sens., vol. 151, pp. 223–236, May 2019.


\bibitem{SMLFR}
Dong Z, Gu Y, Liu T. Generative ConvNet foundation model with sparse modeling and low-frequency reconstruction for remote sensing image interpretation[J]. IEEE Transactions on Geoscience and Remote Sensing, 2024, 62: 1-16.

\bibitem{foundation-model}
Awais M, Naseer M, Khan S, et al. Foundation Models Defining a New Era in Vision: a Survey and Outlook[J]. IEEE Transactions on Pattern Analysis and Machine Intelligence, 2025.

\bibitem{self-supervised}
Liu X, Zhang F, Hou Z, et al. Self-supervised learning: Generative or contrastive[J]. IEEE transactions on knowledge and data engineering, 2021, 35(1): 857-876.

\bibitem{RSP}
Wang D, Zhang J, Du B, et al. An empirical study of remote sensing pre-training[J]. IEEE Transactions on Geoscience and Remote Sensing, 2022, 61: 1-20.

\bibitem{ringmo}
Sun X, Wang P, Lu W, et al. RingMo: A remote sensing foundation model with masked image modeling[J]. IEEE Transactions on Geoscience and Remote Sensing, 2022, 61: 1-22.


\bibitem{pis}
An X, He W, Zou J, et al. pre-train a Remote Sensing Foundation Model by Promoting Intra-Instance Similarity[J]. IEEE Transactions on Geoscience and Remote Sensing, 2024.

\bibitem{fgmae}
Wang Y, Hernández H H, Albrecht C M, et al. Feature guided masked autoencoder for self-supervised learning in remote sensing[J]. IEEE Journal of Selected Topics in Applied Earth Observations and Remote Sensing, 2024.





\bibitem{AGCAM}
Chen L, Cai X, Li Z, et al. Where is my attention? An explainable AI exploration in water detection from SAR imagery[J]. International Journal of Applied Earth Observation and Geoinformation, 2024, 130: 103878.

\bibitem{RCC-MRF}
Zhang A, Yang X, Fang S, et al. Region level SAR image classification using deep features and spatial constraints[J]. ISPRS journal of photogrammetry and remote sensing, 2020, 163: 36-48.

\bibitem{HCRF}
Liang W, Wu Y, Li M, et al. High-resolution SAR image classification using context-aware encoder network and hybrid conditional random field model[J]. IEEE Transactions on Geoscience and Remote Sensing, 2020, 58(8): 5317-5335.

\bibitem{CCNR}
Wu Z, Hou B, Guo X, et al. CCNR: Cross-regional context and noise regularization for SAR image segmentation[J]. International Journal of Applied Earth Observation and Geoinformation, 2023, 121: 103363.

\bibitem{MCANet}
Li X, Zhang G, Cui H, et al. MCANet: A joint semantic segmentation framework of optical and SAR images for land use classification[J]. International Journal of Applied Earth Observation and Geoinformation, 2022, 106: 102638.

\bibitem{circular module}
Li W, Sun K, Li W, et al. Aligning semantic distribution in fusing optical and SAR images for land use classification[J]. ISPRS Journal of Photogrammetry and Remote Sensing, 2023, 199: 272-288.

\bibitem{SoftFormer}
Liu R, Ling J, Zhang H. SoftFormer: SAR-optical fusion transformer for urban land use and land cover classification[J]. ISPRS Journal of Photogrammetry and Remote Sensing, 2024, 218: 277-293.

\bibitem{position}
C. Doersch, A. Gupta, and A. A. Efros, “Unsupervised visual representation learning by context prediction,” in Proc. IEEE Int. Conf. Comput. Vis. (ICCV), Dec. 2015, pp. 1422–1430.


\bibitem{jigsaw}
M. Noroozi and P. Favaro, “Unsupervised learning of visual representations by solving jigsaw puzzles,” in Proc. Eur. Conf. Comput. Vis. Cham, Switzerland: Springer, 2016, pp. 69–84.

\bibitem{inpainting}
R. Zhang, P. Isola, and A. A. Efros, “Colorful image colorization,” in Proc. Eur. Conf. Comput. Vis. Cham, Switzerland: Springer, 2016, pp. 649–666.

\bibitem{JPSSL}
Ren Z, Lu Y, Hou B, et al. JPSSL: SAR Terrain Classification Based on Jigsaw Puzzles and FC-CRF[J]. Remote Sensing, 2024, 16(9): 1635.

\bibitem{RotANet}
Wen Z, Liu Z, Zhang S, et al. Rotation awareness based self-supervised learning for SAR target recognition with limited training samples[J]. IEEE Transactions on Image Processing, 2021, 30: 7266-7279.

\bibitem{MAE}
He K, Chen X, Xie S, et al. Masked autoencoders are scalable vision learners[C]//Proceedings of the IEEE/CVF conference on computer vision and pattern recognition. 2022: 16000-16009.

\bibitem{SimMIM}
Xie Z, Zhang Z, Cao Y, et al. Simmim: A simple framework for masked image modeling[C]//Proceedings of the IEEE/CVF conference on computer vision and pattern recognition. 2022: 9653-9663.

\bibitem{S2FL}
Xue Z, Yu X, Yu A, et al. Self-supervised feature learning for multimodal remote sensing image land cover classification[J]. IEEE Transactions on Geoscience and Remote Sensing, 2022, 60: 1-15.

\bibitem{SAR-JEPA}
Li W, Yang W, Liu T, et al. Predicting gradient is better: Exploring self-supervised learning for SAR ATR with a joint-embedding predictive architecture[J]. ISPRS Journal of Photogrammetry and Remote Sensing, 2024, 218: 326-338.

\bibitem{MSFA}
Y. Li, X. Li, W. Li, Q. Hou, L. Liu, M.-M. Cheng, and J. Yang, “SARDet-100K: Towards open-source benchmark and toolkit for large scale SAR object detection,” in Proc. Adv. Neural Inf. Process. Syst. (NeurIPS), 2024.

\bibitem{GLCNet}
Li H, Li Y, Zhang G, et al. Global and local contrastive self-supervised learning for semantic segmentation of HR remote sensing images[J]. IEEE Transactions on Geoscience and Remote Sensing, 2022, 60: 1-14.

\bibitem{IndexNet}
Muhtar D, Zhang X, Xiao P. Index your position: A novel self-supervised learning method for remote sensing images semantic segmentation[J]. IEEE Transactions on Geoscience and Remote Sensing, 2022, 60: 1-11.

\bibitem{CSSL}
Yang M, Jiao L, Liu F, et al. Coarse-to-fine contrastive self-supervised feature learning for land-cover classification in SAR images with limited labeled data[J]. IEEE Transactions on Image Processing, 2022, 31: 6502-6516.


\bibitem{a billion scale}
K. Cha, J. Seo and T. Lee, ``A Billion-scale Foundation Model for Remote Sensing Images," in IEEE Journal of Selected Topics in Applied Earth Observations and Remote Sensing, doi: 10.1109/JSTARS.2024.3401772.


\bibitem{SatMAE}
Cong Y, Khanna S, Meng C, et al. Satmae: Pre-training transformers for temporal and multi-spectral satellite imagery[J]. Advances in Neural Information Processing Systems, 2022, 35: 197-211.


\bibitem{RVSA}
Wang D, Zhang Q, Xu Y, et al. Advancing plain vision transformer toward remote sensing foundation model[J]. IEEE Transactions on Geoscience and Remote Sensing, 2022, 61: 1-15.




\bibitem{Scale-MAE}
Reed C J, Gupta R, Li S, et al. Scale-mae: A scale-aware masked autoencoder for multiscale geospatial representation learning[C]//Proceedings of the IEEE/CVF International Conference on Computer Vision. 2023: 4088-4099.




\bibitem{SARATR-X}
Li W, Yang W, Hou Y, et al. SARATR-X: Towards Building A Foundation Model for SAR Target Recognition[J]. IEEE Transactions on Image Processing, 2025.

\bibitem{SkySense}
X. Guo, J. Lao, B. Dang, Y. Zhang, L. Yu, L. Ru, L. Zhong, Z. Huang, K. Wu, D. Hu et al., “SkySense: A multi-modal remote sensing foundation model towards universal interpretation for earth observation imagery,” in Proc. IEEE Comput. Soc. Conf. Comput. Vis. Pattern Recognit. (CVPR), 2024, pp. 27672–27683.

\bibitem{OFA-Net}
Xiong Z, Wang Y, Zhang F, et al. One for all: Toward unified foundation models for Earth vision[C]//IGARSS 2024-2024 IEEE International Geoscience and Remote Sensing Symposium. IEEE, 2024: 2734-2738.

\bibitem{SkySenseV2}
Zhang Y, Ru L, Wu K, et al. Skysense v2: A unified foundation model for multi-modal remote sensing[C]//Proceedings of the IEEE/CVF International Conference on Computer Vision. 2025: 9136-9146.

\bibitem{RingMoE}
Bi H, Feng Y, Tong B, et al. RingMoE: Mixture-of-Modality-Experts Multi-Modal Foundation Models for Universal Remote Sensing Image Interpretation[J]. arXiv preprint arXiv:2504.03166, 2025.

\bibitem{Geography-Aware SSL}
Ayush K, Uzkent B, Meng C, et al. Geography-aware self-supervised learning[C]//Proceedings of the IEEE/CVF International Conference on Computer Vision. 2021: 10181-10190.


\bibitem{SeCo}
Manas O, Lacoste A, Giró-i-Nieto X, et al. Seasonal contrast: Unsupervised pre-training from uncurated remote sensing data[C]//Proceedings of the IEEE/CVF International Conference on Computer Vision. 2021: 9414-9423.

\bibitem{PIS}
An X, He W, Zou J, et al. pre-train a Remote Sensing Foundation Model by Promoting Intra-Instance Similarity[J]. IEEE Transactions on Geoscience and Remote Sensing, 2024.

\bibitem{CMID}
Muhtar D, Zhang X, Xiao P, et al. Cmid: A unified self-supervised learning framework for remote sensing image understanding[J]. IEEE Transactions on Geoscience and Remote Sensing, 2023, 61: 1-17.

\bibitem{instance1}
Z. Wu, Y. Xiong, S. X. Yu, and D. Lin, “Unsupervised feature learning via non-parametric instance discrimination,” in Proc. IEEE/CVF Conf. Comput. Vis. Pattern Recognit., Jun. 2018, pp. 3733–3742.

\bibitem{instance2}
P. Bachman, R. D. Hjelm, and W. Buchwalter, “Learning representations by maximizing mutual information across views,” in Proc. Adv. Neural Inf. Process. Syst., vol. 32, 2019, pp. 1–12.

\bibitem{instance3}
M. Ye, X. Zhang, P. C. Yuen, and S.-F. Chang, “Unsupervised embedding learning via invariant and spreading instance feature,” in Proc. IEEE/CVF Conf. Comput. Vis. Pattern Recognit. (CVPR), Jun. 2019, pp. 6203–6212.


\bibitem{simclr}
T. Chen, S. Kornblith, M. Norouzi, and G. Hinton, “A simple framework for contrastive learning of visual representations,” in Proc. Int. Conf. Mach. Learn., 2020, pp. 1597–1607.

\bibitem{moco}
He K, Fan H, Wu Y, et al. Momentum contrast for unsupervised visual representation learning[C]//Proceedings of the IEEE/CVF conference on computer vision and pattern recognition. 2020: 9729-9738.

\bibitem{mocov2}
Chen X, Fan H, Girshick R, et al. Improved baselines with momentum contrastive learning[J]. arXiv preprint arXiv:2003.04297, 2020.

\bibitem{byol}
Grill J B, Strub F, Altché F, et al. Bootstrap your own latent-a new approach to self-supervised learning[J]. Advances in neural information processing systems, 2020, 33: 21271-21284.

\bibitem{simsiam}
X. Chen and K. He, “Exploring simple Siamese representation learning,” in Proc. IEEE/CVF Conf. Comput. Vis. Pattern Recognit. (CVPR), Jun. 2021, pp. 15745–15753.


\bibitem{barlowtwins}
Zbontar J, Jing L, Misra I, et al. Barlow twins: Self-supervised learning via redundancy reduction[C]//International conference on machine learning. PMLR, 2021: 12310-12320.

\bibitem{mocov3}
Chen X, Xie S, He K. An empirical study of training self-supervised vision transformers[C]//Proceedings of the IEEE/CVF international conference on computer vision. 2021: 9640-9649.

\bibitem{dino}
Caron M, Touvron H, Misra I, et al. Emerging properties in self-supervised vision transformers[C]//Proceedings of the IEEE/CVF international conference on computer vision. 2021: 9650-9660.

\bibitem{pixpro}
Xie Z, Lin Y, Zhang Z, et al. Propagate yourself: Exploring pixel-level consistency for unsupervised visual representation learning[C]//Proceedings of the IEEE/CVF conference on computer vision and pattern recognition. 2021: 16684-16693.

\bibitem{resim}
Xiao T, Reed C J, Wang X, et al. Region similarity representation learning[C]//Proceedings of the IEEE/CVF International Conference on Computer Vision. 2021: 10539-10548.

\bibitem{ssccl}
Dong Z, Liu T, Gu Y. Spatial and semantic consistency contrastive learning for self-supervised semantic segmentation of remote sensing images[J]. IEEE Transactions on Geoscience and Remote Sensing, 2023.

\bibitem{scrl}
Roh B, Shin W, Kim I, et al. Spatially consistent representation learning[C]//Proceedings of the IEEE/CVF conference on computer vision and pattern recognition. 2021: 1144-1153.




\bibitem{pytorch}
A. Paszke et al., “PyTorch: An imperative style, high-performance deep learning library,” in Proc. Adv. Neural Inf. Process. Syst., vol. 32, 2019.

\bibitem{resnet}
K. He, X. Zhang, S. Ren, and J. Sun, “Deep residual learning for image recognition,” in Proc. IEEE Conf. Comput. Vis. Pattern Recognit. (CVPR), Jun. 2016, pp. 770–778.


\bibitem{deeplab}
L.-C. Chen, Y. Zhu, G. Papandreou, F. Schroff, and H. Adam, “Encoder–decoder with Atrous separable convolution for semantic image segmentation,” in Proc. Eur. Conf. Comput. Vis. (ECCV), 2018, pp. 801–818.

\bibitem{DANet}
Fu J, Liu J, Tian H, et al. Dual attention network for scene segmentation[C]//Proceedings of the IEEE/CVF conference on computer vision and pattern recognition. 2019: 3146-3154.

\bibitem{PSPNet}
Zhao H, Shi J, Qi X, et al. Pyramid scene parsing network[C]//Proceedings of the IEEE conference on computer vision and pattern recognition. 2017: 2881-2890.

\bibitem{UPerNet}
Xiao T, Liu Y, Zhou B, et al. Unified perceptual parsing for scene understanding[C]//Proceedings of the European conference on computer vision (ECCV). 2018: 418-434.

\bibitem{FarSeg}
Zheng Z, Zhong Y, Wang J, et al. Foreground-aware relation network for geospatial object segmentation in high spatial resolution remote sensing imagery[C]//Proceedings of the IEEE/CVF conference on computer vision and pattern recognition. 2020: 4096-4105.


\bibitem{ConvNeXt}
Liu Z, Mao H, Wu C Y, et al. A convnet for the 2020s[C]//Proceedings of the IEEE/CVF conference on computer vision and pattern recognition. 2022: 11976-11986.


\bibitem{Dice}
Milletari F, Navab N, Ahmadi S A. V-net: Fully convolutional neural networks for volumetric medical image segmentation[C]//2016 fourth international conference on 3D vision (3DV). Ieee, 2016: 565-571.




\bibitem{WVResU-Net}
Jamali A, Roy S K, Beni L H, et al. Residual wave vision U-Net for flood mapping using dual polarization Sentinel-1 SAR imagery[J]. International Journal of Applied Earth Observation and Geoinformation, 2024, 127: 103662.

\bibitem{DeepAqua}
Peña F J, Hübinger C, Payberah A H, et al. DeepAqua: Semantic segmentation of wetland water surfaces with SAR imagery using deep neural networks without manually annotated data[J]. International Journal of Applied Earth Observation and Geoinformation, 2024, 126: 103624.


\bibitem{DINOv2}
Oquab M, Darcet T, Moutakanni T, Vo H V, et al. DINOV2: Learning Robust Visual Features without Supervision[J]. Transactions on Machine Learning Research (TMLR), 2024.

\bibitem{ViT}
Dosovitskiy, L. Beyer, A. Kolesnikov, et al. “An Image is Worth 16x16 Words: Transformers for Image Recognition at Scale,” in Proc. Int. Conf. Learning Representations, 2021.


\bibitem{of_add}
Guo W, Li S, Yang J. Scattering prompt tuning: A fine-tuned foundation model for SAR object recognition [C]//Proceedings of the IEEE/CVF Conference on Computer Vision and Pattern Recognition. 2024: 3056-3065.
\end{thebibliography}
\end{document}


%
\title{
  \textit{Supplementary Materials for} \\
  \upshape DI3CL: Contrastive Learning With Dynamic Instances and Contour Consistency for SAR Land-Cover Classification Foundation Model
}

\author{}

\markboth{Journal of \LaTeX\ Class Files,~Vol.~3, No.~6, June~2025}%
{Shell \MakeLowercase{\textit{et al.}}: Bare Demo of IEEEtran.cls for Journals}
%



\maketitle

%
\IEEEpeerreviewmaketitle

\section{EXPERIMENTS}

\subsection{SAR Land-Cover Mapping}

\subsubsection{Full-Image Inference}
To evaluate the effectiveness of our model in practical land-cover mapping scenarios, we perform full-scene inference on two SAR images. These images are acquired by the Gaofen-3 satellite, with a spatial resolution of 3 meters and HH polarization mode. The scenes are captured in Yan'an and Xianyang, Shaanxi Province, with sizes of $14,240\times12,963$ and $14,952 \times 12,578$ pixels, respectively. 

\begin{figure*}[htbp]
    \centering
    \includegraphics[width=0.85\textwidth]{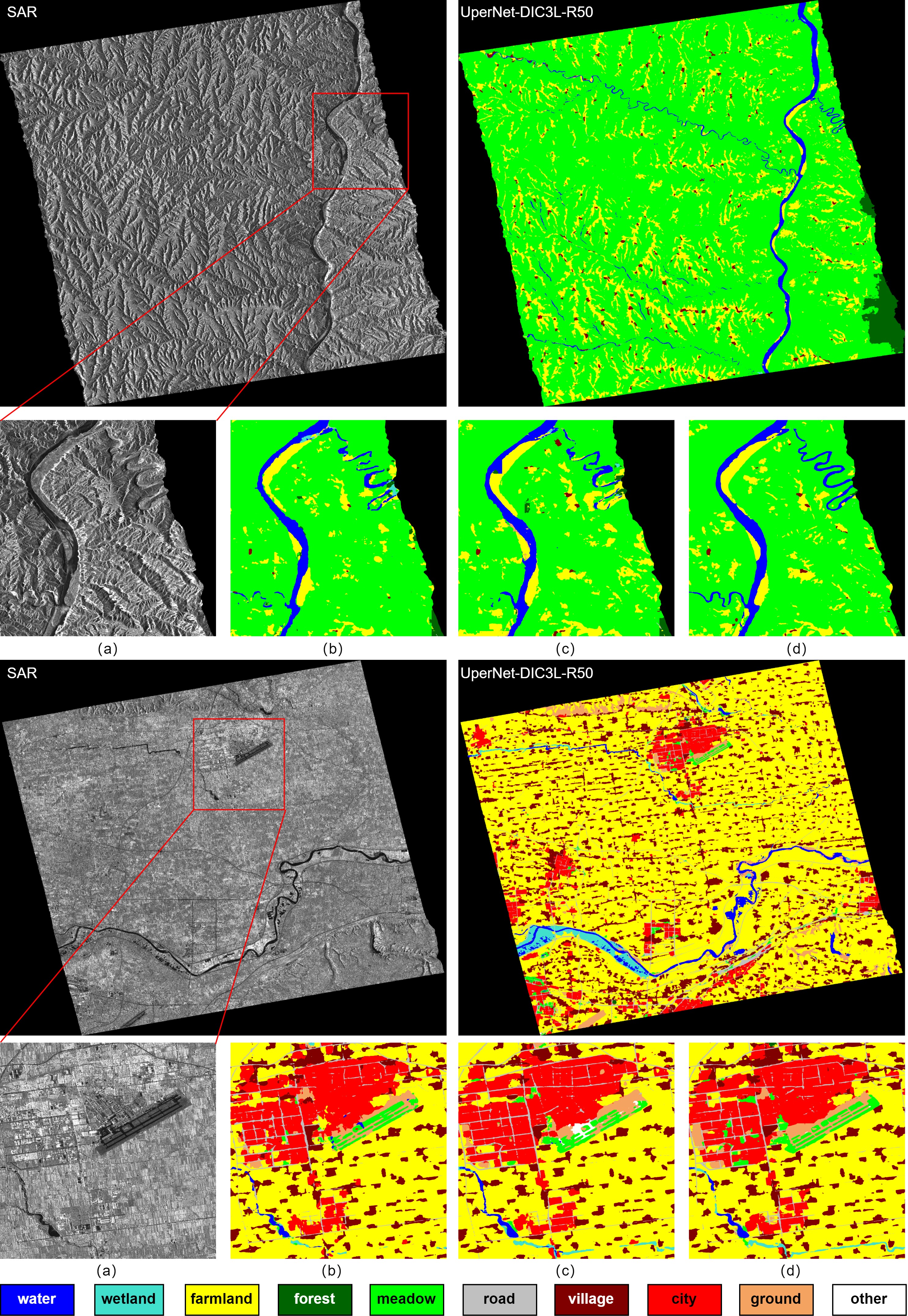} 
    \caption{Visual comparison of full-scene inference results for the SAR-land cover mapping task. The main image presents the output of UperNet-DI3CL-R50, while (a) shows a zoomed-in SAR patch, and (b)–(d) show the corresponding predictions of UperNet, UperNet-SMLFR, and UperNet-DI3CL-R50, respectively.}
    \label{figure_whole_mapping}
\end{figure*}

To mitigate block artifacts during inference, we adopt an overlapping sliding window strategy with a stride of 100 pixels, cropping each scene into $512\times512$ image patches for individual prediction. The final results are reconstructed by stitching the predicted patches. Visualization results are shown in Fig.\ref{figure_whole_mapping}. Our model achieves accurate classification performance and produces visually meaningful results. In particular, in the zoomed-in detail regions, the model demonstrates superior classification accuracy, highlighting its strong generalization capability.

\subsection{SAR Water Detection}

\subsubsection{Full-Image Inference}
\begin{figure*}[htbp]
    \centering
    \includegraphics[width=0.82\textwidth]{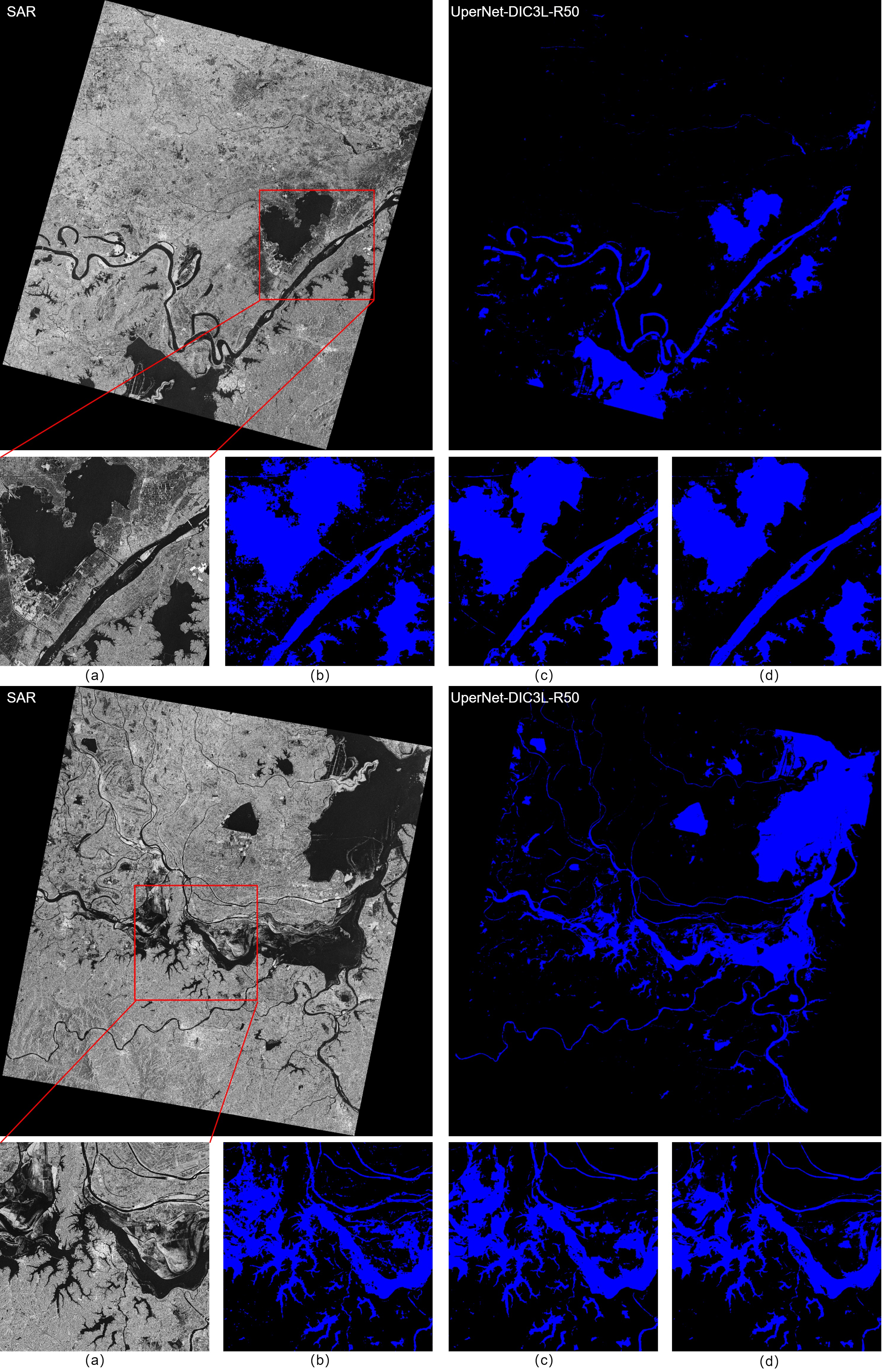} 
    \caption{Visual comparison of full-scene inference results for the SAR water detection task. The main image presents the output of UperNet-DI3CL-R50, while (a) shows a zoomed-in SAR patch, and (b)–(d) show the corresponding predictions of UperNet, UperNet-SMLFR, and UperNet-DI3CL-R50, respectively.}
    \label{figure_whole_water}
\end{figure*}

Similarly, we conduct full-scene inference for water detection on two SAR images, both captured by the Gaofen-3 satellite with a spatial resolution of 5 meters and HV polarization. The images are acquired in the vicinity of Dongting Lake, Hunan Province, with sizes of $32,437\times35,402$ and $30,343\times33,755$ pixels, respectively. The inference is performed using the same configuration as in the full-scene inference for SAR land cover mapping. Fig.\ref{figure_whole_water} presents the visualization results. Our model achieved relatively complete water detection results. As shown in the zoomed-in regions, the detected water boundaries align more closely with the actual scenes, exhibiting fewer spurious noise pixels and smoother contours. However, the model shows limited performance in detecting small water bodies. Nevertheless, by introducing a dedicated water detection head, the pre-trained backbone weights from our model can still serve as a strong foundation for specialized water detection tasks.

\subsection{SAR Road Extraction}

\subsubsection{Full-Image Inference}
Finally, we perform full-scene inference for road extraction on two SAR images, both acquired by the Gaofen-3 satellite with a spatial resolution of 3 meters and HH polarization. These images are captured in Weinan and Xianyang, Shaanxi Province, with sizes of $24,053\times21,792$ and $14,145\times12,986$ pixels, respectively. The inference process followed the same settings as in previous full-scene inference experiments.

\begin{figure*}[htbp]
    \centering
    \includegraphics[width=0.85\textwidth]{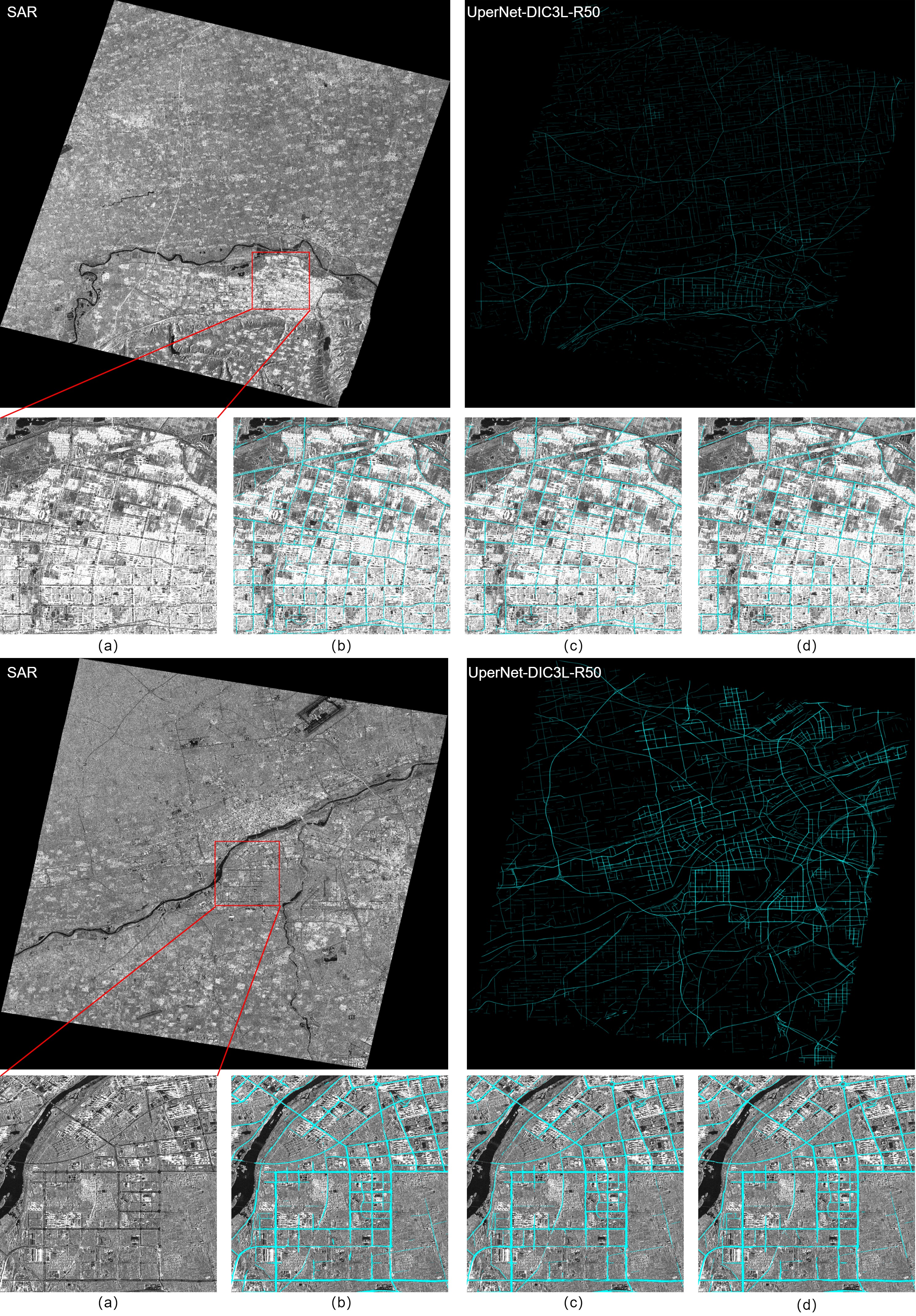} 
    \caption{Visual comparison of full-scene inference results for the SAR road extraction task. The main image presents the output of UperNet-DI3CL-R50, while (a) shows a zoomed-in SAR patch, and (b)–(d) show the corresponding predictions of UperNet, UperNet-RSP, and UperNet-DI3CL-R50, respectively.}
    \label{figure_whole_road}
\end{figure*}

The visualization results are shown in Fig.\ref{figure_whole_road}. Our model achieved relatively continuous road extraction, successfully delineating the basic road network structure. In the zoomed-in areas, the results show higher recall, with more complete and connected road extractions. However, some disconnections and missed extractions are still present. It is worth noting that this experiment focuses on fair comparison with baseline models. We believe that by incorporating a dedicated road extraction head and appropriate post-processing strategies, these limitations can be effectively addressed.

\subsection{Visual Analysis and Spatial Accuracy Comparison}
\begin{figure*}[htbp]
    \centering
    \includegraphics[width=0.7\textwidth]{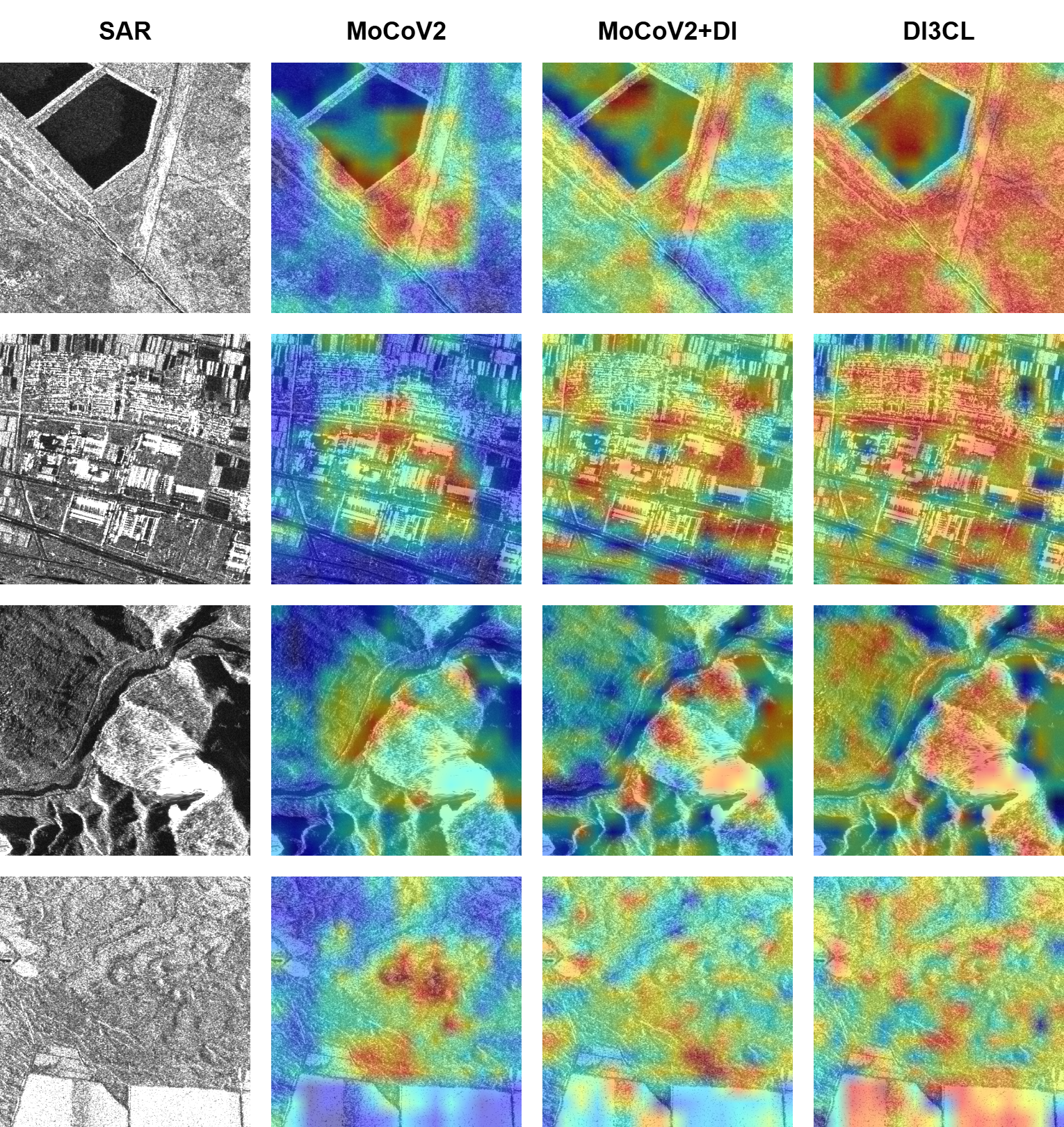} 
    \caption{Qualitative comparison of attention maps using different pre-training strategies on SAR imagery. (a) Baseline (e.g., MoCoV2) exhibits center-biased attention; (b) Integration of the Dynamic Instance (DI) module enables global contextual awareness; (c) Combined with Contour Consistency (CC), the model achieves precise alignment with the structural contours of land-cover objects.}
    \label{attention_map}
\end{figure*}

To qualitatively validate the effectiveness of the proposed Dynamic Instance (DI) and Contour Consistency (CC) modules, we provide additional attention map visualizations using A-GCAM. As illustrated in Fig. \ref{attention_map}, models pre-trained with traditional contrastive learning strategies, such as MoCoV2, tend to focus primarily on the image center while neglecting critical land-cover information in surrounding areas. With the integration of the DI module, the model's attention successfully expands to the global context of the entire scene. However, at this stage, the attention remains relatively scattered and overlooks the unique geometric contours and structural information inherent in SAR imagery. By further incorporating the CC module, the model’s attention not only maintains a global perspective but also precisely aligns with the physical contours of the SAR objects. This synergy ensures that the framework captures both comprehensive contextual awareness and fine-grained structural details.

\ifCLASSOPTIONcaptionsoff
  \newpage
\fi